\documentclass{article}

\PassOptionsToPackage{numbers, sort}{natbib}

\usepackage[preprint]{neurips_2026}
\usepackage[utf8]{inputenc} 
\usepackage[T1]{fontenc}    
\usepackage{microtype}      

\usepackage{amsmath}
\usepackage{amssymb}
\usepackage{amsfonts}       
\usepackage{nicefrac}       

\usepackage[table]{xcolor}
\usepackage{fontawesome5}
\definecolor{crimson}{rgb}{0.86, 0.08, 0.24}
\usepackage{float}
\usepackage{placeins}
\usepackage{hyperref}
\usepackage{url}            

\usepackage{booktabs}       
\usepackage{tabularx}
\usepackage{array}
\usepackage{ragged2e}
\newcolumntype{Y}{>{\RaggedRight\arraybackslash}X}

\usepackage{pifont}
\newcommand{\cmark}{\textcolor{green!60!black}{\ding{51}}} 
\newcommand{\xmark}{\textcolor{red!70!black}{\ding{55}}}   
\usepackage{titletoc}
\hypersetup{
  colorlinks=true,
  citecolor=blue!70,
  linkcolor=red!70,
  urlcolor=blue
}
\usepackage[most]{tcolorbox}

\newtcolorbox{promptbox}[1]{
  enhanced,
  breakable,
  colback=gray!3,
  colframe=black!55,
  boxrule=0.45pt,
  arc=2pt,
  left=6pt,
  right=6pt,
  top=6pt,
  bottom=6pt,
  title=#1,
  fonttitle=\bfseries,
  coltitle=black,
  attach boxed title to top left={xshift=4pt,yshift=-2pt},
  boxed title style={
    colback=gray!15,
    colframe=black!55,
    boxrule=0.45pt,
    arc=2pt
  }
}

\title{OpenHalDet: A Unified Benchmark for Hallucination Detection across Diverse Generation Scenarios}

\author{
\begin{tabular}{c}
\textbf{
Xinyi Li$^{1}$,
Zhen Fang$^{1}$,
Yongxin Deng$^{1}$,
Jinyuan Luo$^{1}$,
Hongnan Ma$^{3}$
} \\
\textbf{
Changdae Oh$^{2}$,
Zijing Shi$^{1}$,
Shanshan Ye$^{1}$,
Hanchen Wang$^{1}$,
Shu-Lin Chen$^{1}$
} \\
\textbf{
Yadan Luo$^{4}$,
Mengyue Yang$^{3}$,
Sean Du$^{5}$,
Sharon Li$^{2}$,
Ling Chen$^{1}$
} \\[5pt]
{\normalfont\mdseries\small
$^{1}$University of Technology Sydney \quad
$^{2}$University of Wisconsin--Madison \quad
$^{3}$University of Bristol
} \\
{\normalfont\mdseries\small
$^{4}$The University of Queensland \quad
$^{5}$Nanyang Technological University
} \\[4pt]
{\normalfont\mdseries\small
\texttt{xinyi.li-1@student.uts.edu.au} \quad
\texttt{zhen.fang@uts.edu.au}
}
\end{tabular}
}

\usepackage[most]{tcolorbox}
\usepackage{enumitem}

\usepackage{booktabs}
\usepackage{longtable}
\usepackage{array}

\begin{document}

\maketitle

\begin{abstract}
Hallucination detection is essential for the reliable deployment of large language models (LLMs). 
However, existing evaluations face two core challenges: inconsistent inference configuration and evaluation, and limited coverage of downstream domains and tasks. 
Consequently, reported detector performance is often difficult to compare, reproduce, and generalize beyond specific experimental settings. 
We introduce \textit{OpenHalDet}, a unified benchmark for hallucination detection across diverse generation scenarios. 
OpenHalDet standardizes the evaluation pipeline, from prompt construction and response generation to truthfulness annotation, detector scoring, and metric computation. 
It supports heterogeneous detector families under different access settings, including black-box methods that use only generated outputs, gray-box methods that rely on probability-based signals, and white-box methods that exploit internal model signals. 
By bringing diverse tasks, models, and detectors into a shared framework, OpenHalDet enables controlled comparison and provides a systematic view of how different detection paradigms behave in LLM applications. 
We release OpenHalDet as an open and extensible codebase to facilitate reproducible evaluation and future development of hallucination detection methods. The code and datasets are available at \url{https://github.com/Nellie179/Hallucination-Detection}.

\end{abstract}

\section{Introduction}

Large language models (LLMs) have demonstrated strong potential across a wide range of real-world applications~\cite{zhao2026surveylargelanguagemodels,LitianLiu}. 
Despite these advances, hallucination in generated text remains a critical challenge, where LLMs produce outputs that are grammatically and logically coherent but lack factual accuracy or verifiable evidence~\cite{park2025steer,triviaqa,truthfulqa}. 
This issue is particularly consequential in high-risk domains such as healthcare, law, and finance~\cite{harvey2025don,roustan2025clinicians,kang2023deficiency}, where the dissemination of incorrect information may lead to serious consequences. 
{Unfortunately, LLMs trained via likelihood-based next-token prediction are inherently prone to hallucination~\cite{maynez2020faithfulness,kalai2024calibrated,banerjee2025llms,kalai2025language}, and zero-hallucination guarantees are unattainable in general without additional structural assumptions \cite{10.1145/3618260.3649777, DBLP:conf/stoc/KalavasisMV25}. 
Therefore, effectively detecting hallucinations in model outputs, known as \textit{hallucination detection}~\cite{ren2022out,li2023halueval,manakul2023selfcheckgpt,kuhn2023semantic}, has become a promising research direction for improving the reliability and safety of LLMs.}

Prior work \cite{10.1145/3703155,zhang-etal-2025-sirens} has proposed a wide range of hallucination detection methods, which can be broadly grouped according to their model-access requirements into black-box, gray-box, and white-box methods. 
1) \textit{Black-box methods} rely only on externally observable token-space signals, such as consistency patterns across generated outputs~\cite{Lin2022TeachingMT,manakul2023selfcheckgpt,lin2024generating}. 
2) \textit{Gray-box methods} further exploit probability-based signals exposed during generation, including token probabilities, sequence likelihoods, or entropy-based uncertainty measures~\cite{ren2023outofdistribution,malinin2021uncertainty,duan-etal-2024-shifting}.
3) \textit{White-box methods} require access to model-internal information, such as hidden states, attention maps, or other intermediate representations~\cite{chen2024inside,DBLP:conf/iclr/BurnsYKS23,azaria2023the}. 
These lines of work have shown promising results and enabled a range of applications, including selective prediction~\cite{yadkori2024mitigating}, domain-specific safety solutions~\cite{asgari2025framework}, reliable retrieval-augmented generation systems~\cite{sun2025redeep}, and agentic inference guardrails~\cite{noel2026spectral}.
However, despite substantial methodological progress, the current evaluation paradigm for hallucination detection still exhibits two flaws that undermine genuine progress: 
1) \textit{inconsistent evaluation and inference configurations} \cite{Janiak2025TheIO,bang-etal-2025-hallulens}, and 
2) \textit{narrowly scoped downstream domains and tasks} \cite{emery2025hallumix}.

Specifically, the first limitation concerns the evaluation protocol: existing studies~\citep{azaria2023the, manakul2023selfcheckgpt, zhang-etal-2025-icr} often adopt different model backbones, prompting methods, decoding strategies, and inference hyperparameters to validate their proposed detectors (\textit{see Appendix~\ref{app:prior_detector_settings}}). Recent work has shown that detector effectiveness can vary substantially across evaluation configurations~\citep{duan-etal-2024-shifting}. 
Therefore, the lack of standardized evaluation and inference configurations hinders apples-to-apples comparisons across methods~\cite{Janiak2025TheIO,bang-etal-2025-hallulens,hallucination-detection}. The second limitation concerns the coverage of evaluation scenarios: most detectors are evaluated on only a small number of datasets covering limited downstream domains and task formats~\cite{emery2025hallumix, urlana-etal-2025-hallucounter}. Since detection strategies may behave differently across domains, tasks, context lengths, and answer formats, such narrow evaluation scope makes it difficult to assess whether reported effectiveness can generalize beyond the specific settings considered in each study.

To address these challenges, we introduce \textit{OpenHalDet}, a unified benchmark with an accompanying well-structured and extensible codebase for hallucination detection. OpenHalDet standardizes the main stages of evaluation, including prompt construction, response generation, truthfulness annotation, detector scoring, and metric reporting, enabling fair comparison across detectors under a shared protocol. It covers \textit{17 datasets}, full evaluations on \textit{4 backbone LLMs}, and selected 70B-scale experiments across diverse generation scenarios (\textit{see Appendix~\ref{app:more_results}}). OpenHalDet further integrates representative black-box, gray-box, and white-box detectors into a common framework, supporting controlled comparison under unified settings. Together, it provides a fair, reproducible, and comprehensive view of hallucination detection across practical LLM application scenarios. We summarize our contributions as follows:

\begin{tcolorbox}[colback=gray!3,colframe=gray!35,boxrule=0.4pt,arc=1mm,left=1mm,right=1mm,top=1mm,bottom=1mm]
\begin{itemize}[leftmargin=*]
    \item \textbf{Comprehensive Hallucination Detection Benchmarks.} We provide a benchmark covering 17 datasets across diverse LLM application scenarios. OpenHalDet groups datasets by scenario, enabling systematic evaluation of hallucination detectors under different task formats.
    \item \textbf{Unified Comparison Across Access Regimes.} We integrate 16 representative detectors with different access assumptions into a shared evaluation interface. This enables fair comparison across methods under the same generation, annotation, and scoring protocol.

\item 
\textbf{A Unified Codebase for Hallucination Detection.} We provide an extensible open-source codebase, {OpenHalDet}, with standardized modules for prompt construction, generation, annotation, detector scoring, and metric reporting. 
It supports heterogeneous datasets and detectors through a shared interface, reducing the engineering effort for reproducible evaluation.

\item 
\textbf{Insights.} Our evaluation yields three findings: 
1) detector effectiveness is scenario- and backbone-dependent, with self-reported confidence varying strongly across settings; 
2) richer model access raises the performance ceiling, but does not guarantee robust gains; 
3) evidence acquisition often dominates cost, making accuracy-only comparisons incomplete.
\end{itemize}
\end{tcolorbox}
\vspace{-0.7em}

\section{Task Formulation, Benchmark Scope, and Dataset Construction}
\label{sec:benchmark}

Here, we first introduce the task formulation studied in our benchmark. 
We then describe the scope of our benchmark.
Further, we describe how datasets with different original formats are converted into a unified evaluation format. 
Finally, we present the evaluation metrics used in our experiments.

\vspace{-0.7em}
\subsection{Task Formulation}
\label{subsec:task_definition}

Let $\mathcal{Q}$ and $\mathcal{A}$ be the spaces of inputs and outputs. Following \cite{du2024haloscope,park2025steer,deng2026beyond}, we consider a base LLM as a probability distribution $\mathbb{P}_{\boldsymbol{\theta}}(\cdot)$ over token sequences, where $\boldsymbol{\theta}$ denotes the model parameters.
Given an input token sequence $\mathbf{Q}=[q_1,\ldots,q_k] \in \mathcal{Q}$, where $q_j$ is the $j$-th token, the LLM generates an output sequence $\mathbf{A}=[q_{k+1},\ldots,q_{k+a}] \in \mathcal{A}$ of length $a$ by sampling from
\begin{equation}
\mathbb{P}_{\boldsymbol{\theta}}(\mathbf{A} \mid \mathbf{Q}) =
\prod_{j=k+1}^{k+a} \mathbb{P}_{\boldsymbol{\theta}}(q_j \mid q_{<j}),
\label{eq:llm_autoreg}
\end{equation}
where $q_{<j} = (q_1,\ldots,q_{j-1})$ denotes the preceding context. Given the underlying \emph{truthful-response domain} $\mathbb{P}_{Q,T}$, 
which is a joint distribution over $\mathcal{Q}\times \mathcal{A}$, 
the goal of hallucination detection is to learn a detector $G$ such that, 
for any input-output pair $(\mathbf{Q},\mathbf{A})$ with 
$\mathbf{Q}\sim \mathbb{P}_{Q}$ and 
$\mathbf{A}\sim \mathbb{P}_{\boldsymbol{\theta}}(\cdot \mid \mathbf{Q})$,
\begin{equation}\label{Def:HD}
G(\mathbf{Q},\mathbf{A}) =
0,~\text{if } \mathbf{A}\text{ is semantically aligned with } \mathbf{T}\sim\mathbb{P}_{T|Q}(\cdot \mid \mathbf{Q});~\text{otherwise},~G(\mathbf{Q},\mathbf{A}) =
1,
\end{equation}
where $0$ indicates that $\mathbf{A}$ is a truthful response and $1$ indicates that $\mathbf{A}$ is a hallucinated response. 

Note that Eq.~\eqref{Def:HD} corresponds to the standard response-level hallucination detection setting, where the detector judges the truthfulness of the entire generated output. 
Recent studies have also considered finer-grained variants, including entity-level \cite{yeh2025halluentity}, atomic-fact-level \cite{min2023factscore}, and sentence-level \cite{manakul2023selfcheckgpt} hallucination detection, which identify hallucinations at the level of entities, atomic facts, or individual sentences, respectively. 
Finer-grained settings can also support response-level detection, since the truthfulness of a response may also depend on the correctness of its constituent entities, facts, and sentences. 
Moreover, response-level hallucination detection remains the most widely studied setting. 
Therefore, this work mainly focuses on the response-level hallucination detection.

\begin{table*}[t]
\centering
\scriptsize
\caption{
\textbf{Dataset coverage of OpenHalDet across diverse generation scenarios.}
The format column specifies the task structure used for unified prompt construction and annotation.
The last two columns indicate whether prior hallucination benchmarks explicitly cover each broad scenario.
}
\label{tab:datasets}
\setlength{\tabcolsep}{1.5pt}
\renewcommand{\arraystretch}{0.96}
\begin{tabular}{
>{\raggedright\arraybackslash}p{3.3cm}
>{\raggedright\arraybackslash}p{4.8cm}
>{\raggedright\arraybackslash}p{2.6cm}
>{\centering\arraybackslash}p{1.5cm}
>{\centering\arraybackslash}p{1.5cm}
}
\toprule
\textbf{Scenario} 
& \textbf{Datasets} 
& \textbf{Format} 
& \textbf{HaluEval~\citep{li2023halueval}} 
& \textbf{HalluMix~\citep{emery2025hallumix}} \\
\midrule

QA: Multiple-choice 
& ARC-Challenge~\citep{Clark2018ThinkYH}; CommonsenseQA~\citep{talmor-etal-2019-commonsenseqa} 
& MCQ 
& \xmark 
& \xmark \\

QA: Open-ended 
& TriviaQA~\citep{triviaqa}; TruthfulQA~\citep{truthfulqa} 
& Short answer 
& \cmark 
& \cmark \\

QA: Reading comprehension 
& SQuAD~v2~\citep{rajpurkar-etal-2018-know} 
& Span extraction 
& \xmark 
& \xmark \\

QA: Multi-hop 
& HotpotQA~\citep{yang-etal-2018-hotpotqa} 
& Cross-document reasoning 
& \cmark 
& \xmark \\

QA: Conversational 
& CoQA~\citep{reddy-etal-2019-coqa} 
& Dialogue 
& \xmark 
& \xmark \\

QA: Grounded 
& HaluEval-QA~\citep{li2023halueval} 
& Context-based QA 
& \cmark 
& \cmark \\

\midrule

Retrieval-augmented generation 
& RAGTruth~\citep{niu-etal-2024-ragtruth} 
& RAG generation 
& \xmark 
& \cmark \\

Summarization 
& XSum~\citep{narayan-etal-2018-dont} 
& Abstractive summarization 
& \cmark 
& \cmark \\

Mathematical reasoning 
& GSM8K~\citep{Cobbe2021TrainingVT}; SVAMP~\citep{patel-etal-2021-nlp} 
& Chain-of-thought 
& \xmark 
& \xmark \\

Scientific reasoning 
& TheoremQA~\citep{chen-etal-2023-theoremqa} 
& Chain-of-thought 
& \xmark 
& \xmark \\

Code generation 
& HumanEval~\citep{Chen2021EvaluatingLL}; MBPP~\citep{Austin2021ProgramSW} 
& Code synthesis 
& \xmark 
& \xmark \\

Agentic tasks 
& xLAM-Agent~\citep{zhang-etal-2025-xlam} 
& Tool invocation 
& \xmark 
& \xmark \\

Multilingual evaluation 
& Belebele~\citep{bandarkar-etal-2024-belebele} 
& Multilingual MCQ 
& \xmark 
& \xmark \\

\bottomrule
\end{tabular}

\vspace{0.2em}
\begin{minipage}{0.90\textwidth}
\scriptsize
\textit{Note.} A check mark indicates explicit coverage of the corresponding broad scenario, not necessarily the same dataset or exact sub-format.
HalluMix is marked for RAG due to its multi-document grounded generation setting.
\end{minipage}
\vspace{-0.9em}
\end{table*}

\subsection{Benchmark Scope: Unified, Scenario-Aware, And Model-Diverse Evaluation}
\label{subsec:datasets}
\textbf{Towards Unified and Fair Evaluation.}
Existing hallucination detection studies differ in model choices, task settings, data construction pipelines, and evaluation protocols (\textit{see Appendix~\ref{app:prior_detector_settings}}), making cross-study comparisons difficult. 
OpenHalDet addresses this fragmentation by consolidating datasets, LLM-generated responses, representative detectors, and standardized metrics under a common evaluation protocol. 
This unified framework enables fair comparison across detector families and supports reproducible extension of hallucination detection methods.


\textbf{Reliable Evaluation Across Diverse Scenarios.} Many hallucination detection studies~\citep{park2025steer, du2024haloscope, zhang-etal-2025-icr, azaria2023the, wang2025latent} evaluate detectors on a narrow set of tasks, typically factual question answering~\citep{truthfulqa, triviaqa} or mathematical reasoning~\citep{Cobbe2021TrainingVT, patel-etal-2021-nlp}. 
This limited coverage makes it difficult to assess generalization across broader real-world scenarios. 
OpenHalDet addresses this gap with a scenario-aware evaluation suite, summarized in Table~\ref{tab:datasets}. 
We organize datasets by application scenario rather than as a flat collection, enabling systematic analysis of detector accuracy and robustness across task settings.

\textbf{Fine-Grained Coverage Across QA Settings.}
Question answering (QA) is widely used for hallucination detection, but it is not a homogeneous task: QA settings vary in answer format, evidence availability, and reasoning demand. 
OpenHalDet therefore separates QA into multiple-choice, open-ended, reading-comprehension, multi-hop, conversational, and context-grounded settings. 
ARC-Challenge~\citep{Clark2018ThinkYH} and CommonsenseQA~\citep{talmor-etal-2019-commonsenseqa} cover multiple-choice science and commonsense reasoning; TriviaQA~\citep{triviaqa} and TruthfulQA~\citep{truthfulqa} cover open-ended factual and truthfulness-oriented QA; SQuAD~v2~\citep{rajpurkar-etal-2018-know} and HotpotQA~\citep{yang-etal-2018-hotpotqa} evaluate evidence-grounded answer extraction, including unanswerable and multi-hop cases. 
CoQA~\citep{reddy-etal-2019-coqa} adds conversational context, while HaluEval-QA~\citep{li2023halueval} provides QA-style hallucination examples with human annotations. 
This design enables fine-grained analysis of detector reliability across QA formats, grounding conditions, and reasoning requirements.

\textbf{Broad Coverage Beyond Question Answering.}
OpenHalDet further extends evaluation beyond QA to retrieval-augmented generation, summarization, reasoning, code generation, agentic tool use, and multilingual understanding. 
This scope captures settings where detection must account for evidence support, summarization faithfulness, multi-step reasoning, code correctness, tool-call validity, and cross-lingual comprehension. 
RAGTruth~\citep{niu-etal-2024-ragtruth} and XSum~\citep{narayan-etal-2018-dont} cover evidence-grounded generation and abstractive summarization; GSM8K~\citep{Cobbe2021TrainingVT} and TheoremQA~\citep{chen-etal-2023-theoremqa} cover mathematical and theorem-based reasoning; HumanEval~\citep{Chen2021EvaluatingLL} and MBPP~\citep{Austin2021ProgramSW} cover code generation. 
We further include xLAM-Agent~\citep{zhang-etal-2025-xlam} for tool use and Belebele~\citep{bandarkar-etal-2024-belebele} for multilingual evaluation. 
This design enables analysis of detector reliability across a wider range of generation contexts.


\textbf{Model-Diverse Evaluation.}
OpenHalDet evaluates detectors across recent open-weight LLMs from the Llama and Qwen families, reflecting the need for robustness beyond a single LLM. 
Since many prior studies rely on earlier or single backbones, it remains unclear whether their conclusions transfer to newer models. 
The selected backbones are representative recent open-weight LLMs from two widely used model families. Appendix~\ref{app:LLMs} summarizes the selected models together with representative Hugging Face download statistics, providing a practical indicator of community adoption and supporting our controlled analysis of detector stability across model families.

\subsection{Datasets Collection Protocol}
\label{subsec:collection}

\textbf{Challenges.} Constructing a comprehensive hallucination detection benchmark requires more than simply aggregating datasets from different sources. 
\textit{Three} major challenges arise from dataset heterogeneity. 
First, \textit{different task types exhibit different input structures}, including choice-based inputs, external evidence, programming constraints, and tool specifications. 
Second, \textit{the expected response formats vary considerably}, spanning short answers, option labels, long-form text, reasoning traces, code, and structured tool calls. 
Third, \textit{the collected datasets provide reference answers in different forms} (\textit{see Appendix~\ref{app:datasets}}), from a single gold answer to multiple acceptable answers. 
\textit{These differences make it difficult to evaluate all datasets directly within a single unified pipeline}.

\textbf{Unified Instance Schema.}
OpenHalDet uses a unified instance schema as the core abstraction of the benchmark pipeline. 
Instead of flattening heterogeneous datasets into raw prompts, each instance is decomposed into shared semantic roles:
\[
s = (t, i, c, q, \omega, r^{+}, r^{-}),
\]
where \(t\) denotes the task type, \(i\) the task instruction, \(c\) optional context or constraints, \(q\) the primary input, \(\omega\) task-specific options or constraints, and \(r^{+}\) and \(r^{-}\) denote correct and known incorrect references. 
The task type \(t\) determines both the prompt rule and the expected response format.

This schema preserves task-specific structure while providing a common interface across datasets. 
Input components such as passages, retrieved evidence, answer options, tool specifications, or testing constraints are stored in role-specific fields rather than collapsed into an unstructured prompt. 
Reference information is normalized through \(r^{+}\) and \(r^{-}\), allowing the same pipeline to support single gold answers, multiple acceptable answers, and known incorrect answers. 
Given \(t\), OpenHalDet renders only the relevant fields into the model input, standardizing prompt construction while preserving task-specific information. 
\textit{Details and examples are provided in Appendix~\ref{app:datasets}}.

\textbf{Label Assignment.}
To obtain comparable labels across heterogeneous tasks, we use a unified response-level annotation protocol with GPT-4o-mini, following the broader practice of LLM-based evaluation~\citep{10.5555/3666122.3668142}. 
Given the task input, optional context, reference answers, known incorrect answers when available, and the candidate response, the annotator assigns one of three labels: \texttt{correct}, \texttt{hallucination}, or \texttt{abstention}. 
We use response-level labels rather than token-, span-, sentence-, or claim-level annotations, since finer units are difficult to define consistently across QA, summarization, reasoning, code, tool-use, and multilingual settings. 
For binary detector evaluation, \texttt{hallucination} is treated as the positive class and \texttt{correct} as the negative class; abstentions and invalid annotations are excluded. 
\textit{More details are provided in Appendix~\ref{app:annotation}}.

\vspace{-0.5em}
\subsection{Evaluation Metrics}
\label{subsec:metrics}

Following~\citep{park2025steer, du2024haloscope}, we use \textit{AUROC} as the primary metric. 
Each detector assigns a scalar hallucination score to each response, where higher scores indicate higher hallucination risk. 
AUROC evaluates whether hallucinated responses receive higher scores than correct responses across decision thresholds. 
In addition to the main benchmark tables, we provide a separate supplementary cost analysis to characterize detector efficiency rather than to define the primary ranking. 
We report \textit{Cost@N}, the wall-clock time for applying a detector to $N$ samples under a fixed hardware and evaluation protocol, decomposed into feature preparation, training, and inference time. 
\textit{See more details in Appendix~\ref{app:cost_analysis}}.

\begin{figure*}[t]
    \centering
    \includegraphics[width=\textwidth,height=0.2\textheight]    {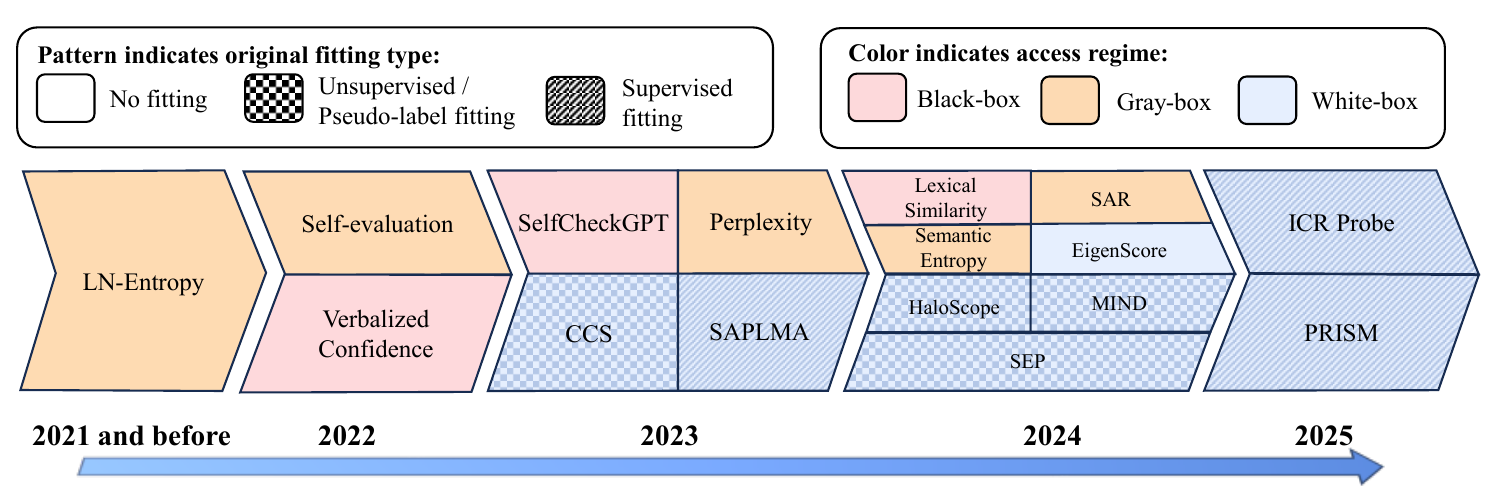}
    \caption{
    Timeline and taxonomy of hallucination detection methods supported by OpenHalDet.
    Methods are organized by publication year.
    Colors indicate model-access regimes, including black-box, gray-box, and white-box detectors.
    Fill patterns indicate whether the method requires no additional training, calibration or fitting, or supervised adaptation. See Table \ref{tab:baselines} for more details.
    }
    \label{fig:method_timeline}
\end{figure*}

\section{Supported Detection Methods and  Evaluation Protocol}
\label{sec:methods}
This section provides a concise overview of the 16 hallucination detection methods shown in Figure~\ref{fig:method_timeline}. We focus on representative methods with publicly available implementations or sufficiently clear algorithmic descriptions for reproducible implementation, \textit{see more details in Appendix~\ref{app:detector_impl}}. 


\subsection{Black-box Detectors}
\label{subsec:black_box}
Black-box detectors perform hallucination detection using only text outputs, without access to token probabilities, sequence likelihoods, hidden states, or attention maps. 
They therefore suit closed-source or API-only cases, where users can query the model but cannot inspect its internal signals.

We categorize them by the type of textual signal used: \emph{verbalized-confidence methods}, which rely on a self-reported confidence statement from a single output, and \emph{sample-consistency methods}, which estimate hallucination risk from consistency across multiple sampled responses. 1) Verbalized-confidence methods are a small but practical family of black-box detectors that use the model's self-reported confidence as a hallucination signal. 
Such methods are lightweight and easy to deploy, but their reliability depends on whether the model's stated confidence is calibrated with factual correctness. 
We include \textsc{Verbalized Confidence}~\citep{Lin2022TeachingMT} as a representative baseline for this category. 2) Sample-consistency methods estimate hallucination risk by generating multiple responses to the same input and measuring disagreement among sampled outputs. 
\textsc{SelfCheckGPT}~\citep{manakul2023selfcheckgpt} checks whether independently sampled responses support or contradict the main response, whereas \textsc{Lexical Similarity}~\citep{lin2024generating} captures surface-level agreement across sampled outputs. 

\vspace{-0.4em}
\subsection{Gray-box Detectors}
\label{subsec:gray_box}
Gray-box detectors rely on probability-based signals exposed during generation, including token probabilities, sequence likelihoods, and related generation scores. They require more model access than black-box detectors, but do not use hidden states, attention maps, or other internal representations.
In OpenHalDet, gray-box detectors are represented by likelihood-based uncertainty methods.

These methods estimate hallucination risk from probability-based signals associated with the model output, including token probabilities, sequence likelihoods, and uncertainty over sampled generations.
We organize them into two groups: \emph{single-output likelihood methods} and \emph{sampling-based uncertainty methods}. 1) For {single-output likelihood} methods, \textsc{Perplexity}~\citep{ren2023outofdistribution} scores a generated output using token-level likelihoods, while \textsc{Self-evaluation}~\citep{Kadavath2022LanguageM} asks the model to assess whether its own output is true or false; following our score orientation, we use the probability of the ``False'' label, equivalently one minus the probability of correctness, as the hallucination score. 2) For {sampling-based uncertainty} methods, \textsc{LN-Entropy}~\citep{malinin2021uncertainty} measures length-normalized sequence-level uncertainty from sampled outputs and their likelihoods.
\textsc{SAR}~\citep{duan-etal-2024-shifting} further reweights this uncertainty by semantic relevance among samples; for scalability, we use its sentence-level variant.
\textsc{Semantic Entropy}~\citep{kuhn2023semantic} clusters sampled outputs into semantic equivalence classes and computes uncertainty over the resulting semantic distribution.
All methods in this group are training-free, but require access to probability or likelihood information from the target model.

\vspace{-0.4em}
\subsection{White-box Detectors}
\label{subsec:white_box}

White-box detectors require access to internal model signals, such as hidden states, attention maps, layerwise activations, or derived internal features. 
In contrast to black-box and gray-box methods, they directly exploit internal representations rather than relying only on observable text or probability-based generation signals. 
Such information may provide richer evidence for hallucination detection, but imposes the strongest access assumptions. 
As a result, white-box detectors are typically more applicable to open-source or fully accessible models than to closed-source API-only systems.

These methods use internal representations extracted from the target LLM to detect hallucination. 
We organize them according to how internal signals are used: \emph{representation-consistency scores}, \emph{contrastive or subspace-based objectives}, \emph{hidden-state probes}, and \emph{prompt- or dynamics-guided internal features}. 
For \emph{representation-consistency} methods, \textsc{EigenScore}~\citep{chen2024inside} measures representation-level consistency across stochastic generations by analyzing the spectral geometry of hidden-state representations. 
For \emph{contrastive or subspace-based} methods, \textsc{CCS}~\citep{DBLP:conf/iclr/BurnsYKS23} learns a contrast-consistent direction from paired hidden states, while \textsc{HaloScope}~\citep{du2024haloscope} estimates hallucination-related membership in a latent representation subspace and trains a truthfulness classifier from the estimated memberships. 
For \emph{hidden-state probe} methods, \textsc{SAPLMA}~\citep{azaria2023the} trains a supervised classifier on input-output hidden states to predict statement truthfulness. 
\textsc{MIND}~\citep{su-etal-2024-unsupervised} uses generation-time hidden states in an unsupervised training framework for real-time hallucination detection, and \textsc{SEP}~\citep{kossen2025semantic} trains probes to approximate semantic entropy from hidden states of a single generation. 
For \emph{prompt- or dynamics-guided internal features}, \textsc{PRISM}~\citep{zhang-etal-2025-prompt} uses prompt-guided hidden states to improve cross-domain supervised detection, whereas \textsc{ICR Probe}~\citep{zhang-etal-2025-icr} constructs ICR scores from hidden-state updates and attention maps to track cross-layer residual-stream dynamics.

\definecolor{bbrow}{RGB}{249,236,233}
\definecolor{gbrow}{RGB}{246,238,222}
\definecolor{wbrow}{RGB}{229,238,240}
\definecolor{famavg}{RGB}{221,221,221}
\definecolor{modelgray}{RGB}{215,215,215}

\providecommand{\na}{--}

\begin{table*}[t]
\centering
\scriptsize
\caption{
AUROC results (\%) aggregated by scenario across backbone LLMs.
Columns follow the task taxonomy in Table~\ref{tab:datasets}: 
QA averages ARC-Challenge, CommonsenseQA, TriviaQA, TruthfulQA, SQuAD\_V2, HotpotQA, CoQA, and HaluEval-QA when available; 
RAG, Sum., Sci., Agent, and Multi. report RAGTruth, XSum, TheoremQA, xLAM-Agent, and Belebele, respectively; 
Math averages GSM8K and SVAMP; 
and Code averages HumanEval and MBPP.
Higher values are better. SelfCheck-BERT and SelfCheck-NLI denote the BERTScore- and NLI-based SelfCheckGPT.
}
\label{tab:scenario_backbone_auroc}

\setlength{\tabcolsep}{1.0pt}
\renewcommand{\arraystretch}{1.03}

\begin{tabular}{@{}l*{16}{c}@{}}
\toprule
&
\multicolumn{2}{c}{\textbf{QA}} &
\multicolumn{2}{c}{\textbf{RAG}} &
\multicolumn{2}{c}{\textbf{Sum.}} &
\multicolumn{2}{c}{\textbf{Math}} &
\multicolumn{2}{c}{\textbf{Sci.}} &
\multicolumn{2}{c}{\textbf{Code}} &
\multicolumn{2}{c}{\textbf{Agent}} &
\multicolumn{2}{c}{\textbf{Multi.}} \\
\cmidrule(lr){2-3}
\cmidrule(lr){4-5}
\cmidrule(lr){6-7}
\cmidrule(lr){8-9}
\cmidrule(lr){10-11}
\cmidrule(lr){12-13}
\cmidrule(lr){14-15}
\cmidrule(lr){16-17}

\rowcolor{modelgray}
\multicolumn{17}{@{}l}{\textbf{\texttt{Llama}}} \\
\textbf{Method} &
\textbf{3.1-8B} & \textbf{3.2-3B} &
\textbf{3.1-8B} & \textbf{3.2-3B} &
\textbf{3.1-8B} & \textbf{3.2-3B} &
\textbf{3.1-8B} & \textbf{3.2-3B} &
\textbf{3.1-8B} & \textbf{3.2-3B} &
\textbf{3.1-8B} & \textbf{3.2-3B} &
\textbf{3.1-8B} & \textbf{3.2-3B} &
\textbf{3.1-8B} & \textbf{3.2-3B} \\
\midrule

\rowcolor{bbrow}
Verbalized Conf. & 68.46 & 63.00 & 61.23 & 60.09 & 55.91 & 51.05 & 66.40 & 55.07 & 63.49 & 50.96 & 58.76 & 53.95 & 54.46 & 50.34 & 51.09 & 51.38 \\
\rowcolor{bbrow}
SelfCheck-BERT & 72.91 & 71.20 & 69.98 & 54.33 & 61.50 & 61.27 & 73.36 & 67.45 & 73.92 & 74.07 & 67.35 & 68.32 & 78.82 & 70.27 & 73.84 & 76.27 \\
\rowcolor{bbrow}
SelfCheck-NLI & 73.69 & 69.80 & 71.25 & 60.57 & 64.37 & 69.98 & 80.97 & 76.64 & 52.55 & 55.82 & 55.44 & 56.21 & 66.49 & 68.25 & 66.07 & 74.89 \\
\rowcolor{bbrow}
Lexical Sim. & 65.69 & 71.02 & 59.91 & 57.34 & 56.79 & 60.74 & 74.68 & 68.82 & 63.80 & 66.64 & 62.88 & 71.91 & 62.59 & 66.05 & 65.77 & 75.37 \\
\rowcolor{famavg}
\textit{Black-box Avg.} & 70.19 & 68.76 & 65.59 & 58.08 & 59.64 & 60.76 & 73.85 & 67.00 & 63.44 & 61.87 & 61.11 & 62.60 & 65.59 & 63.73 & 64.19 & 69.48 \\

\rowcolor{gbrow}
Perplexity & 75.99 & 71.24 & 70.76 & 62.07 & 64.35 & 57.91 & 78.34 & 66.62 & 64.48 & 51.30 & 63.78 & 54.35 & 83.24 & 65.80 & 69.21 & 79.07 \\
\rowcolor{gbrow}
Self-eval. & 71.55 & 68.46 & 72.95 & 59.22 & 63.62 & 59.34 & 79.84 & 75.84 & 65.92 & 57.69 & 68.67 & 77.40 & 62.86 & 59.56 & 50.46 & 52.24 \\
\rowcolor{gbrow}
LN-Entropy & 66.07 & 69.71 & 68.45 & 54.33 & 54.57 & 55.22 & 75.99 & 72.99 & 73.35 & 66.01 & 71.01 & 67.85 & 67.41 & 69.29 & 70.43 & 78.87 \\
\rowcolor{gbrow}
SAR & 65.67 & 69.98 & 67.33 & 52.34 & 53.12 & 53.90 & 76.18 & 71.99 & 73.84 & 62.07 & 68.38 & 69.34 & 68.32 & 70.49 & 71.95 & 80.27 \\
\rowcolor{gbrow}
Semantic Ent. & 62.50 & 65.96 & 51.26 & 60.34 & 52.13 & 56.05 & 50.90 & 51.64 & 61.30 & 52.68 & 66.63 & 57.70 & 60.23 & 63.24 & 64.78 & 76.36 \\
\rowcolor{famavg}
\textit{Gray-box Avg.} & 68.36 & 69.07 & 66.15 & 57.66 & 57.56 & 56.48 & 72.25 & 67.82 & 67.78 & 57.95 & 67.69 & 65.33 & 68.41 & 65.68 & 65.37 & 73.36 \\

\rowcolor{wbrow}
EigenScore & 62.02 & 64.48 & 56.45 & 54.77 & 50.76 & 53.28 & 72.14 & 66.48 & 51.25 & 62.48 & 72.14 & 77.39 & 62.19 & 65.37 & 53.00 & 66.66 \\
\rowcolor{wbrow}
CCS & 55.51 & 53.82 & 54.43 & 57.21 & 54.08 & 54.78 & 68.97 & 64.58 & 74.77 & 61.74 & 53.58 & 59.66 & 52.63 & 50.93 & 50.78 & 59.68 \\
\rowcolor{wbrow}
HaloScope & 60.57 & 60.89 & 61.28 & 57.74 & 51.67 & 53.99 & 63.95 & 64.35 & 58.90 & 57.04 & 65.03 & 65.03 & 62.05 & 59.42 & 67.17 & 69.67 \\
\rowcolor{wbrow}
SAPLMA & 81.03 & 76.04 & 77.98 & 70.34 & 66.60 & 56.57 & 83.32 & 76.47 & 69.87 & 78.61 & 71.24 & 73.23 & 93.15 & 85.83 & 68.22 & 67.95 \\
\rowcolor{wbrow}
MIND & 77.52 & 78.40 & 65.23 & 62.91 & 61.79 & 59.41 & 84.08 & 79.00 & 79.23 & 80.95 & 63.95 & 65.53 & 92.18 & 88.03 & 72.68 & 75.11 \\
\rowcolor{wbrow}
SEP & 81.00 & 65.83 & 74.46 & 51.35 & 64.80 & 56.15 & 82.54 & 63.52 & 76.24 & 57.43 & 68.41 & 60.98 & 87.82 & 65.77 & 69.76 & 73.48 \\
\rowcolor{wbrow}
ICR Probe & 60.65 & 57.88 & 57.34 & 53.38 & 58.14 & 61.37 & 56.72 & 53.17 & 73.90 & 54.65 & 65.33 & 62.91 & 61.36 & 71.35 & 58.57 & 62.49 \\
\rowcolor{wbrow}
PRISM & 81.98 & 74.11 & 80.21 & 62.00 & 67.65 & 62.41 & 81.06 & 74.82 & 80.51 & 79.67 & 67.75 & 65.35 & 87.47 & 77.25 & 61.38 & 67.78 \\
\rowcolor{famavg}
\textit{White-box Avg.} & 70.04 & 66.43 & 65.92 & 58.71 & 59.44 & 57.25 & 74.10 & 67.80 & 70.58 & 66.57 & 65.93 & 66.26 & 74.86 & 70.49 & 62.70 & 67.85 \\

\midrule

\rowcolor{modelgray}
\multicolumn{17}{@{}l}{\textbf{\texttt{Qwen3}}} \\
\textbf{Method} &
\textbf{8B} & \textbf{14B} &
\textbf{8B} & \textbf{14B} &
\textbf{8B} & \textbf{14B} &
\textbf{8B} & \textbf{14B} &
\textbf{8B} & \textbf{14B} &
\textbf{8B} & \textbf{14B} &
\textbf{8B} & \textbf{14B} &
\textbf{8B} & \textbf{14B} \\
\midrule

\rowcolor{bbrow}
Verbalized Conf. & 61.94 & 64.44 & 63.52 & 67.82 & 51.79 & 54.23 & 53.86 & 58.89 & 59.17 & 51.28 & 56.55 & 60.52 & 69.36 & 58.94 & 51.68 & 64.55 \\
\rowcolor{bbrow}
SelfCheck-BERT & 62.06 & 60.25 & 53.97 & 54.08 & 57.45 & 54.17 & 56.84 & 66.41 & 59.00 & 64.61 & 53.16 & 58.14 & 67.17 & 70.19 & 54.63 & 52.66 \\
\rowcolor{bbrow}
SelfCheck-NLI & 62.57 & 62.91 & 67.86 & 70.04 & 59.16 & 58.89 & 68.24 & 61.08 & 52.99 & 50.49 & 59.29 & 62.50 & 61.56 & 59.80 & 64.98 & 67.92 \\
\rowcolor{bbrow}
Lexical Sim. & 59.53 & 60.85 & 55.87 & 52.66 & 55.92 & 51.94 & 70.07 & 82.39 & 71.56 & 66.36 & 59.45 & 62.23 & 65.10 & 63.52 & 51.68 & 55.68 \\
\rowcolor{famavg}
\textit{Black-box Avg.} & 61.53 & 62.11 & 60.31 & 61.15 & 56.08 & 54.81 & 62.25 & 67.19 & 60.68 & 58.19 & 57.11 & 60.85 & 65.80 & 63.11 & 55.74 & 60.20 \\

\rowcolor{gbrow}
Perplexity & 66.64 & 67.20 & 68.97 & 73.76 & 55.29 & 54.13 & 63.04 & 77.60 & 64.79 & 55.33 & 57.80 & 67.81 & 78.60 & 79.90 & 59.88 & 73.73 \\
\rowcolor{gbrow}
Self-eval. & 76.85 & 73.91 & 56.66 & 53.99 & 60.21 & 61.57 & 77.25 & 81.90 & 71.40 & 70.92 & 64.50 & 61.55 & 79.32 & 72.09 & 54.18 & 63.34 \\
\rowcolor{gbrow}
LN-Entropy & 64.67 & 62.52 & 69.87 & 73.76 & 55.29 & 50.97 & 65.47 & 82.53 & 74.07 & 69.35 & 57.80 & 58.13 & 78.60 & 68.80 & 59.88 & 55.69 \\
\rowcolor{gbrow}
SAR & 60.90 & 62.02 & 64.32 & 74.65 & 56.04 & 50.63 & 70.17 & 78.36 & 72.87 & 66.13 & 62.33 & 54.84 & 67.85 & 70.74 & 55.75 & 55.45 \\
\rowcolor{gbrow}
Semantic Ent. & 58.96 & 58.68 & 61.23 & 57.89 & 52.32 & 52.01 & 58.26 & 61.55 & 58.21 & 62.33 & 64.43 & 57.04 & 58.30 & 58.49 & 51.69 & 55.69 \\
\rowcolor{famavg}
\textit{Gray-box Avg.} & 65.60 & 64.87 & 64.21 & 66.81 & 55.83 & 53.86 & 66.84 & 76.39 & 68.27 & 64.81 & 61.37 & 59.87 & 72.53 & 70.00 & 56.28 & 60.78 \\

\rowcolor{wbrow}
EigenScore & 59.36 & 59.84 & 55.98 & 53.12 & 54.45 & 50.06 & 68.91 & 81.99 & 68.94 & 66.00 & 62.89 & 53.73 & 60.59 & 68.15 & 56.28 & 54.48 \\
\rowcolor{wbrow}
CCS & 58.24 & 61.70 & 56.87 & 57.45 & 51.24 & 53.70 & 73.39 & 71.78 & 65.80 & 65.56 & 63.03 & 53.62 & 50.91 & 55.13 & 60.70 & 50.74 \\
\rowcolor{wbrow}
HaloScope & 59.23 & 56.64 & 56.63 & 57.45 & 52.71 & 52.45 & 68.68 & 69.86 & 52.94 & 60.60 & 53.13 & 50.94 & 56.84 & 57.11 & 58.71 & 62.37 \\
\rowcolor{wbrow}
SAPLMA & 80.88 & 73.98 & 73.42 & 76.42 & 65.03 & 64.95 & 75.49 & 77.18 & 76.45 & 72.30 & 60.14 & 61.59 & 89.77 & 87.97 & 63.51 & 74.92 \\
\rowcolor{wbrow}
MIND & 75.40 & 78.31 & 79.21 & 83.69 & 67.38 & 66.07 & 73.59 & 79.37 & 79.44 & 80.64 & 62.26 & 54.01 & 88.48 & 87.59 & 72.08 & 75.20 \\
\rowcolor{wbrow}
SEP & 64.82 & 72.90 & 59.21 & 53.90 & 52.77 & 65.09 & 57.66 & 58.30 & 55.12 & 63.81 & 59.35 & 69.38 & 74.28 & 83.13 & 63.92 & 71.30 \\
\rowcolor{wbrow}
ICR Probe & 60.62 & 60.31 & 57.87 & 60.12 & 51.59 & 51.96 & 59.01 & 70.89 & 50.28 & 55.16 & 57.07 & 52.98 & 60.26 & 50.36 & 60.05 & 60.83 \\
\rowcolor{wbrow}
PRISM & 73.68 & 70.97 & 60.34 & 68.97 & 65.52 & 66.50 & 70.45 & 71.88 & 74.53 & 72.09 & 52.12 & 56.35 & 80.74 & 82.62 & 62.53 & 57.37 \\
\rowcolor{famavg}
\textit{White-box Avg.} & 66.53 & 66.83 & 62.44 & 63.89 & 57.59 & 58.85 & 68.40 & 72.66 & 65.44 & 67.02 & 58.75 & 56.58 & 70.23 & 71.51 & 62.22 & 63.40 \\

\bottomrule
\end{tabular}

\vspace{-4.0mm}

\end{table*}

\vspace{-0.7em}
\subsection{Access-Aware Evaluation Protocol}
\label{subsec:unified_eval}

\textbf{Challenges.}
A major obstacle to comprehensive hallucination-detector evaluation is the engineering burden of heterogeneous signal extraction. 
Existing detectors operate under different access levels: black-box methods require generated texts or multiple stochastic samples, gray-box methods require predictive distributions such as token log-probabilities, and white-box methods require internal model signals such as layer-wise hidden states or attention maps. 
In practice, evaluating these methods across new LLMs and datasets often requires method-specific data preparation, signal extraction, and scoring pipelines. 
This fragmentation makes large-scale comparison costly and complicates the reproduction of baselines across heterogeneous tasks and backbone models.

\textbf{Decoupled Evaluation Pipeline.}
To reduce the engineering burden of heterogeneous detector evaluation, OpenHalDet decouples response generation, signal extraction, and detector scoring. 
For each instance, the pipeline first constructs a shared evaluation cache containing the primary generated response, stochastic samples, token log-probabilities, and layer-wise hidden states when available. 
Detectors then operate offline on the subset of cached signals allowed by their access regime: surface texts for black-box methods, probability-based generation signals for gray-box methods, and internal representations for white-box methods. 
For specialized white-box detectors, derived internal features are extracted with standardized routines while keeping the generated response fixed.

This design reduces detector integration to a common data-loading and scoring interface, improves comparability by reusing the same generated responses and stochastic samples across methods, and improves reproducibility by centralizing likelihood and representation extraction. 
It also makes OpenHalDet extensible: new datasets can be added through the unified instance schema, and new detectors can be integrated by specifying their required access signals and scoring function. 
As a result, comparisons are less confounded by method-specific input preparation and more directly reflect detector scoring behavior. 
\textit{More details are provided in Appendix~\ref{app:detector_impl}.}

\section{Experiments}
\label{sec:experiments}
\vspace{-0.6em}
\subsection{Experimental Setup}
\vspace{-0.4em}

We conduct a comprehensive evaluation of OpenHalDet on 17 datasets and four backbone LLMs: Llama-3.1-8B-Instruct, Llama-3.2-3B-Instruct~\citep{DBLP:journals/corr/abs-2407-21783}, Qwen3-8B, and Qwen3-14B~\citep{DBLP:journals/corr/abs-2505-09388}. 
We additionally report experiments on Llama-3.3-70B-Instruct in Appendix~\ref{app:more_results}. 
To ensure fair and reproducible comparison, all detectors are evaluated under a unified protocol for prompt construction, response generation, annotation, detector scoring, and metric computation. 
For sample-based methods, we generate five stochastic responses per input with temperature \(1.0\) and top-\(p=0.9\)~\citep{manakul2023selfcheckgpt, kuhn2023semantic}, and reuse the same sampled responses across all applicable detectors. 
For internal-state methods, we use mean-pooled hidden states over generated answer tokens as the default representation~\citep{du2024haloscope, su-etal-2024-unsupervised, zhang-etal-2025-icr}. 
For each dataset, we create stratified 60/20/20 train/validation/test splits that preserve the hallucinated/correct label ratio.
All experiments use a fixed random seed of 42 and are conducted on NVIDIA H100 GPUs. \textit{More details about method-specific implementation are documented in Appendix~\ref{app:detector_impl}.}

\vspace{-0.7em}
\subsection{Main Results}
Table~\ref{tab:scenario_backbone_auroc} reports AUROC (\%) aggregated by scenario under the task taxonomy in Table~\ref{tab:datasets}. 
Multi-dataset scenarios are averaged over constituent datasets, while single-dataset scenarios are reported directly. Detailed per-dataset results and selected Llama-3.3-70B results are provided in Appendix~\ref{app:more_results}. 
We highlight three findings: the detector's effectiveness depends on scenario and backbone; model access improves the ceiling, but not robustness; and evidence acquisition dominates practical cost.

\textbf{Detector Effectiveness Depends on Scenario and Backbone.}
No detector family uniformly dominates across all reported settings. 
For example, on Llama-3.2-3B-Instruct, gray-box methods obtain the highest overall family average, slightly above black-box and white-box methods, with averages of 66.47, 66.07, and 65.91, respectively. 
The leading family also changes across scenarios: on the same backbone, gray-box methods are strongest on multilingual evaluation with an average of 73.36, while white-box methods are stronger on science with an average of 66.57. 
These shifts show that hallucination detection cannot be characterized by a single dataset, scenario, or target backbone.

\textbf{More Model Access Does Not Guarantee Better Detection.}
Access to logits or hidden states provides additional evidence, but it does not by itself determine detector quality. 
Gray-box methods remain competitive with white-box methods: on Llama-3.2-3B-Instruct, the gray-box family average is slightly higher than the white-box average, 66.47 versus 65.91, while on Qwen3-8B the two are close, 64.61 versus 64.97. 
Within the white-box family, performance also varies substantially; for example, on Llama-3.2-3B-Instruct, \textsc{MIND} reaches 75.45 overall AUROC, while \textsc{CCS} obtains 56.67. 
Thus, detector performance depends not only on the access regime, but also on how the available evidence is selected, modeled, and converted into a hallucination-risk score.

\textbf{Black-box Signals Are Limited but Still Task-dependent.}
Black-box detectors rely only on output text or additional generations, which generally limits their performance relative to methods using likelihoods or internal states. 
At the family level, black-box methods are usually below the strongest access regimes, with overall averages of 60.56 on Qwen3-8B, 66.07 on Llama-3.2-3B-Instruct, 67.64 on Llama-3.1-8B-Instruct, and 61.79 on Qwen3-14B. 
Within the black-box family, direct verbalized confidence is generally weaker than sample-consistency signals. 
For example, on Llama-3.2-3B-Instruct, \textsc{Verbalized Confidence} obtains an overall average of 57.99, while \textsc{SelfCheckGPT-BERTScore}, \textsc{SelfCheckGPT-NLI}, and lexical similarity reach 69.25, 67.86, and 69.16, respectively. 
This suggests that comparing stochastic generations is generally more informative than relying on the model's stated confidence, although the usefulness of such consistency signals still depends on the task and comparison metric.

\begin{figure}[t]
    \centering
    \includegraphics[width=\textwidth]{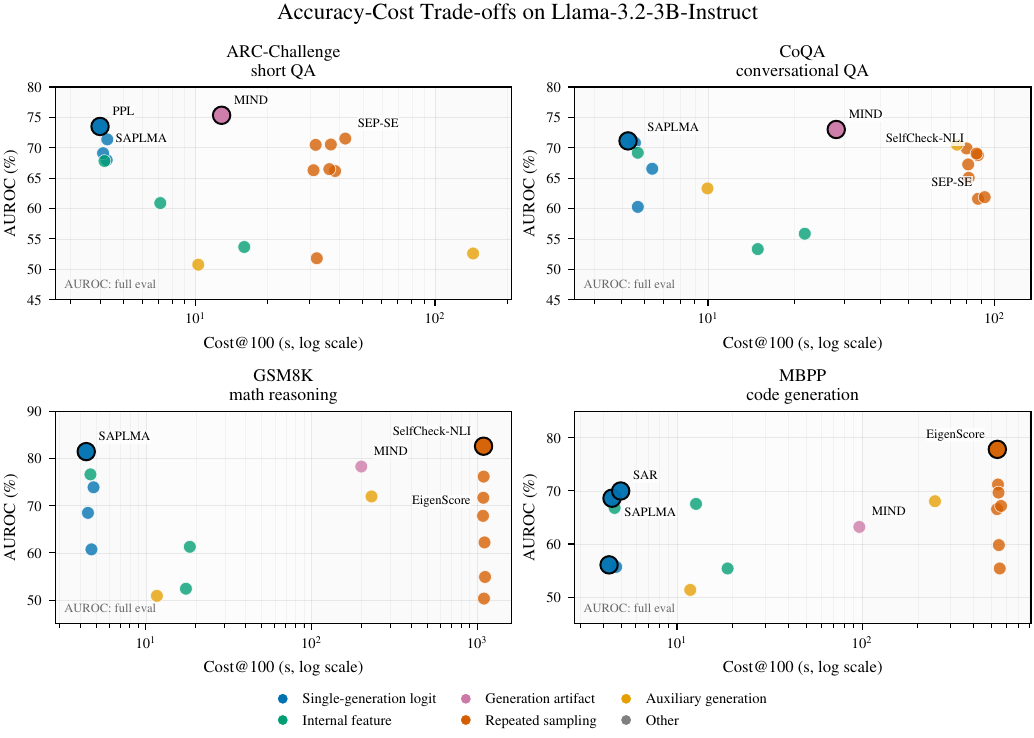}
    \caption{
    Accuracy--cost trade-offs on Llama-3.2-3B-Instruct across representative scenarios.
    Each point denotes a detector. The y-axis reports full-evaluation AUROC, and the x-axis reports artifact-inclusive Cost@100 on a log scale. 
    Colors indicate the evidence-acquisition type. 
    Repeated-generation methods are consistently more expensive, while methods reusing single-generation signals remain low-cost. 
    The benefit of expensive evidence is task-dependent: it is less pronounced on QA-style tasks but becomes more competitive on reasoning and code-generation tasks.
    }
    \label{fig:accuracy_cost_tradeoff}
\end{figure}

\textbf{Accuracy Can Hide Large Cost Differences.}
Figure~\ref{fig:accuracy_cost_tradeoff} provides a cost-aware view on Llama-3.2-3B-Instruct by plotting full-evaluation AUROC against artifact-inclusive Cost@100. 
Detectors with similar AUROC can differ substantially in cost depending on how they acquire evidence. 
Methods that reuse single-generation likelihoods or cached internal features usually remain low-cost, whereas sampling-based methods incur additional cost from repeated generations. 
This extra evidence is not uniformly beneficial: in some scenarios, low-cost likelihood or internal-state methods already approach the performance of sampling-based detectors, while in others repeated-generation signals provide competitive accuracy at higher cost. 
These results show that detector comparisons should report not only accuracy, but also the evidence-acquisition cost required to obtain the score.

Overall, hallucination detection performance depends jointly on the task scenario, target backbone, access regime, and evidence-acquisition strategy. 
White-box methods are often strong, but their gains vary across datasets and models. 
Gray-box methods are competitive practical baselines and can match or exceed weaker internal-state methods without requiring hidden-state access. 
Black-box methods are generally more limited, with sample-consistency signals usually more reliable than direct verbalized confidence. 
These trends motivate a unified benchmark that compares detectors across diverse tasks, backbones, and access regimes under the same protocol.

\section{Comparison with Existing Hallucination Benchmarks}
Existing benchmarks can be broadly divided into two categories. 
The first category focuses on constructing hallucination evaluation datasets or taxonomies. 
HaluEval~\citep{li2023halueval} provides generated and human-annotated hallucinated samples across question answering, dialogue, and summarization; RAGTruth~\citep{niu-etal-2024-ragtruth} offers case-level and word-level hallucination annotations for RAG responses; and HalluLens~\citep{bang-etal-2025-hallulens} further studies hallucination taxonomy and evaluation settings. 
These benchmarks provide data resources, but \textit{their primary goal is to evaluate hallucination behavior or build task-specific hallucination corpora, rather than to systematically compare heterogeneous detector mechanisms}.

The second category is closer to hallucination detector benchmarking. 
HalluMix evaluates several open-source and closed-source hallucination detection systems across domains, task formats, and input settings~\citep{emery2025hallumix}. 
However, \textit{its evaluated systems are mainly deployment-oriented groundedness or API-style detectors}. 
OpenHalDet instead targets academic detector-family evaluation: it covers black-box sample-consistency methods, gray-box likelihood-based uncertainty methods, and white-box internal-state-based methods under a unified protocol. 
In addition, OpenHalDet provides one-command baseline evaluation and a unified detector interface, allowing new hallucination detection methods to be integrated without rebuilding the generation, annotation, scoring, or evaluation pipeline.
\vspace{-1.5em}
\section{Conclusion }
\label{sec:conclusion}
In this paper, we present OpenHalDet, a unified benchmark for evaluating hallucination detection methods across diverse generation scenarios and model-access regimes. 
Using OpenHalDet, we compare 16 representative detectors spanning black-box, gray-box, and white-box settings. 
Our results reveal several key findings: 
1) self-reported confidence is poorly aligned with factual truthfulness; 
2) sample-consistency and likelihood-based uncertainty provide strong practical signals under limited model access; 
3) internal-state methods achieve the highest performance ceiling, but their effectiveness depends on how internal representations are modeled; and 
4) detector performance varies substantially across task scenarios and backbone LLMs. 
Together with standardized protocols for response generation, label assignment, detector scoring, and evaluation, OpenHalDet provides an extensible foundation for fair and reproducible comparison of future hallucination detection methods.



\bibliographystyle{unsrt}
\bibliography{ref}

\newpage
\appendix
\startcontents[appendix]

\section*{Appendix Contents}
\printcontents[appendix]{}{1}{\setcounter{tocdepth}{2}}

\clearpage

\section{Backbone LLMs}
\label{app:LLMs}

\textbf{Backbone selection.} OpenHalDet evaluates hallucination detectors on recent open-weight LLMs from the Llama and Qwen families. 
This choice is motivated by the need to assess detector robustness beyond a single backbone, since detector behavior can vary with model family, scale, instruction tuning, and generation characteristics. 
We use five instruction-tuned backbones spanning two widely adopted open-weight model families and multiple parameter scales: Llama-3.1-8B-Instruct, Llama-3.2-3B-Instruct, Llama-3.3-70B-Instruct, Qwen3-8B, and Qwen3-14B. 
Table~\ref{tab:backbone_llms} summarizes these selected backbones and representative neighboring models from the same families. 
We report Hugging Face download counts as a practical proxy for community adoption, while noting that they are not a definitive measure of model quality, deployment prevalence, or downstream usage.

\begin{table}[h]
\centering
\caption{
Backbone LLMs used in OpenHalDet and representative neighboring open-weight models. 
``Downloads'' denotes Hugging Face downloads in the last month, recorded from the corresponding model pages on May 7, 2026. 
Download counts are dynamic and should be interpreted as a coarse indicator of community adoption rather than a stable benchmark metric.
}
\label{tab:backbone_llms}
\resizebox{\textwidth}{!}{
\begin{tabular}{llrclp{4.8cm}}
\toprule
Model & Family & Params & Downloads & Used in OpenHalDet & Role / Rationale \\
\midrule
\texttt{meta-llama/Llama-3.1-8B-Instruct} 
& Llama & 8B & 9,683,014 & Yes 
& Recent Llama instruction-tuned model; representative mid-size Llama backbone. \\

\texttt{meta-llama/Llama-3.2-3B-Instruct} 
& Llama & 3B & 2,267,285 & Yes 
& Smaller recent Llama backbone, useful for evaluating detector behavior under constrained model scale. \\

\texttt{meta-llama/Llama-3.3-70B-Instruct} 
& Llama & 70B & 778,123 & Yes 
& Large recent Llama instruction-tuned backbone, enabling comparison between smaller and larger Llama models. \\

\texttt{Qwen/Qwen3-8B} 
& Qwen & 8B & 10,813,873 & Yes 
& Recent Qwen3 model with strong community adoption; representative mid-size Qwen backbone. \\

\texttt{Qwen/Qwen3-14B} 
& Qwen & 14B & 2,934,723 & Yes 
& Larger Qwen3 backbone, enabling controlled comparison across Qwen model scales. \\

\midrule
\texttt{meta-llama/Meta-Llama-3-8B-Instruct} 
& Llama & 8B & 1,665,386 & No 
& Earlier Llama-3 instruction-tuned model included as a neighboring release for context. \\

\texttt{Qwen/Qwen2.5-7B-Instruct} 
& Qwen & 7B & 14,206,853 & No 
& Earlier Qwen2.5 instruction-tuned model included as a high-adoption neighboring release. \\

\texttt{Qwen/Qwen2.5-14B-Instruct} 
& Qwen & 14B & 2,642,963 & No 
& Earlier Qwen2.5 model at a similar scale to Qwen3-14B, included for contextual comparison. \\
\bottomrule
\end{tabular}
}
\end{table}

\textbf{Interpretation.} The selected backbones are intended to provide controlled coverage over model family and scale while keeping the benchmark computationally tractable. 
The comparison models in Table~\ref{tab:backbone_llms} are not evaluated in the main experiments; they are included only to contextualize the selected Llama and Qwen backbones relative to nearby open-weight releases. 
Our claims therefore concern detector behavior over the selected Llama and Qwen backbones under the standardized OpenHalDet protocol, rather than universal generalization to all open-weight LLMs.

\section{Prior Detector Settings}
\label{app:prior_detector_settings}

Prior hallucination detectors have been evaluated under diverse experimental settings, with substantial differences in backbone LLMs, task datasets, prompting protocols, and model-access assumptions. 
These differences make cross-study comparisons difficult: a reported performance gap may reflect the detector itself, but it may also arise from the choice of model, task distribution, generation setup, or available evidence. 
Table~\ref{tab:detector_source_setups} summarizes the original evaluation settings associated with the detectors considered in OpenHalDet. 
This comparison is descriptive rather than a re-evaluation of prior work; it motivates the need for a unified protocol in which detectors are compared on the same generated responses, labels, splits, and metric implementations.

\begingroup
\footnotesize
\setlength{\tabcolsep}{5pt}
\renewcommand{\arraystretch}{1.18}

\newcommand{\detrow}[4]{
\texttt{#1} &
\textbf{Backbones.} #2\par
\textbf{Datasets.} #3\par
\textbf{Note.} #4 \\
\addlinespace[0.45em]
}

\begin{longtable}{
>{\raggedright\arraybackslash}p{2.7cm}
>{\raggedright\arraybackslash}p{10.5cm}
}
\caption{
Original backbone LLMs and datasets reported in the source papers or canonical benchmark sources corresponding to the detectors considered in OpenHalDet.
}
\label{tab:detector_source_setups} \\
\toprule
\textbf{Detector} & \textbf{Original evaluation setting} \\
\midrule
\endfirsthead

\toprule
\textbf{Detector} & \textbf{Original evaluation setting} \\
\midrule
\endhead

\bottomrule
\endfoot

\detrow{Selfcheck-BertScore}
{GPT-3}
{WikiBio-based GPT-3 hallucination benchmark}
{Direct detector source.}

\detrow{Selfcheck-NLI}
{GPT-3}
{WikiBio-based GPT-3 hallucination benchmark}
{NLI variant is a later SelfCheckGPT-family instantiation.}

\detrow{Semantic Entropy}
{LLaMA, Falcon, and Mistral families, roughly 7B--70B settings}
{BioASQ, SQuAD, TriviaQA, SVAMP, NQ-Open}
{Direct detector source.}

\detrow{SEP}
{Mistral-7B, Phi-3 Mini/3.8B, Llama-2-7B, Llama-2-70B; long-form analysis also reports Llama-3-70B}
{BioASQ, TriviaQA, NQ Open, SQuAD}
{Direct detector source.}

\detrow{Lexical Similarity}
{OPT-2.7B/6.7B/13B/30B, LLaMA-7B/13B, Vicuna-13B/33B, LLaMA-2-chat-13B, WizardLM-13B}
{CoQA, TriviaQA, SciQ, MedQA, MedMCQA}
{Canonical benchmark instantiation source.}

\detrow{Perplexity}
{OPT-2.7B/6.7B/13B/30B, LLaMA-7B/13B, Vicuna-13B/33B, LLaMA-2-chat-13B, WizardLM-13B}
{CoQA, TriviaQA, SciQ, MedQA, MedMCQA}
{Canonical benchmark instantiation source.}

\detrow{LN-Entropy}
{OPT-2.7B/6.7B/13B/30B, LLaMA-7B/13B, Vicuna-13B/33B, LLaMA-2-chat-13B, WizardLM-13B}
{CoQA, TriviaQA, SciQ, MedQA, MedMCQA}
{Canonical benchmark instantiation source.}

\detrow{Verbalize}
{GPT-3, Vicuna, LLaMA 2, GPT-3.5, GPT-4}
{GSM8K, SVAMP, Date Understanding, Object Counting, StrategyQA, Sports Understanding, Professional Law, Business Ethics}
{Best match for prompt-based verbalized confidence baseline.}

\detrow{Self Evaluation}
{Anthropic LM family, approximately 800M, 2.7B, 12B, and 51B/52B settings}
{TriviaQA, LAMBADA, GSM8K, HumanEval/Codex-style coding tasks, arithmetic tasks; also calibration analyses on BIG-Bench, MMLU, TruthfulQA}
{Canonical self-evaluation / P(True) source.}

\detrow{EigenScore-Internal}
{OPT-6.7B, LLaMA-7B, LLaMA-13B}
{CoQA, SQuAD, NQ}
{Direct detector source.}

\detrow{CCS}
{T5, UnifiedQA, T0, GPT-J, RoBERTa, DeBERTa}
{IMDB, Amazon, AG News, DBpedia, COPA, RTE, BoolQ, QNLI, PIQA, Story Cloze}
{Imported method rather than original hallucination detector.}

\detrow{PRISM}
{Primarily LLaMA2-7B-Chat in the prompt-guided framework experiments}
{Azaria \& Mitchell true/false datasets, including animals, cities, companies, elements, facts, and inventions; also the multi-domain benchmark used in the paper}
{Framework source rather than a single scalar baseline paper.}

\detrow{SAPLMA}
{OPT-6.7B, LLaMA2-7B}
{Cities, Inventions, Chemical Elements, Animals, Companies, Scientific Facts; plus an LLM-generated statements benchmark}
{Direct detector source.}

\detrow{MIND}
{Falcon-40B, GPT-J-6B, LLaMA2-Base-7B, LLaMA2-Chat-7B, LLaMA2-Chat-13B, OPT-7B}
{HELM benchmark; the paper reports hallucination detection over aggregated model outputs from 15 tasks and 700+ datasets}
{Direct detector source.}

\detrow{SAR}
{OPT-2.7B/6.7B/13B/30B, LLaMA-7B/13B, Vicuna-13B/33B, LLaMA-2-chat-13B, WizardLM-13B}
{CoQA, TriviaQA, SciQ, MedQA, MedMCQA}
{Direct detector source.}

\detrow{HaloScope}
{LLaMA-2-chat-7B/13B, OPT-6.7B/13B}
{TruthfulQA, TriviaQA, CoQA, TyDiQA-GP}
{Direct detector source.}

\detrow{ICR\_probe}
{Qwen2.5 family, Llama-3 family; supplementary analysis also includes Gemma-2}
{HaluEval, SQuAD, HotpotQA, TriviaQA}
{Direct detector source.}

\end{longtable}
\endgroup

Overall, prior detector evaluations differ not only in detector implementation but also in backbone models, task distributions, and evidence available to the detector. 
OpenHalDet standardizes these components so that detector comparisons are made on the same generated responses, labels, splits, and metric implementations.

\section{Dataset Processing, Prompt Construction, and Generation Pipeline}
\label{app:datasets}
This section describes the data-processing and response-generation pipeline used by OpenHalDet. 
The goal of the pipeline is to convert heterogeneous task datasets into a shared structured representation, render model-specific prompts in a consistent way, generate LLM responses, and save the intermediate artifacts needed by both black-box and white-box hallucination detectors. 
The automatic annotation protocol is described separately in Appendix~\ref{app:annotation}.

\subsection{Coverage Compared with Representative Hallucination Benchmarks}
\label{app:benchmark_coverage}

Table~\ref{tab:benchmark_comparison} compares OpenHalDet with representative hallucination benchmarks along two axes: scenario coverage and evaluation support. 
Existing benchmarks provide valuable resources for specific settings, such as QA, summarization, or RAG, but they typically focus on a narrower subset of generation scenarios or do not explicitly support systematic detector comparison under different model-access regimes. 
OpenHalDet is designed to complement these resources by unifying heterogeneous tasks under a shared schema and evaluation pipeline, enabling controlled comparison of black-box, gray-box, and white-box hallucination detectors across broader generation settings.

\begin{table*}[h]
\centering
\scriptsize
\caption{
Coverage comparison with representative hallucination benchmarks.
We mark whether each benchmark explicitly covers major generation scenarios and whether it supports systematic hallucination-detector evaluation.
}
\label{tab:benchmark_comparison}
\setlength{\tabcolsep}{2.2pt}
\renewcommand{\arraystretch}{1.08}

\begin{tabular}{lccccccc>{\centering\arraybackslash}p{1.15cm}>{\centering\arraybackslash}p{1.25cm}>{\centering\arraybackslash}p{1.35cm}}
\toprule
\textbf{Benchmark} &
\textbf{QA} &
\textbf{RAG} &
\textbf{Sum.} &
\textbf{Reas.} &
\textbf{Code} &
\textbf{Agent} &
\textbf{Multi.} &
\textbf{Detector Eval.} &
\textbf{Access-aware} &
\textbf{Extensible Pipe.} \\
\midrule
HaluEval~\citep{li2023halueval}
& \cmark & \xmark & \cmark & \xmark & \xmark & \xmark & \xmark & \xmark & \xmark & \xmark \\

RAGTruth~\citep{niu-etal-2024-ragtruth}
& \xmark & \cmark & \xmark & \xmark & \xmark & \xmark & \xmark & \cmark & \xmark & \xmark \\

HalluLens~\citep{bang-etal-2025-hallulens}
& \cmark & \xmark & \cmark & \cmark & \xmark & \xmark & \xmark & \xmark & \xmark & \xmark \\

HalluMix~\citep{emery2025hallumix}
& \cmark & \cmark & \cmark & \cmark & \xmark & \xmark & \xmark & \cmark & \xmark & \xmark \\

\rowcolor{gray!12}
\textbf{OpenHalDet}
& \cmark & \cmark & \cmark & \cmark & \cmark & \cmark & \cmark & \cmark & \cmark & \cmark \\
\bottomrule
\end{tabular}

\vspace{0.3em}
\begin{minipage}{0.96\textwidth}
\footnotesize
\textit{Note.} ``Detector Eval.'' indicates systematic evaluation of hallucination detection systems or methods.
``Access-aware'' indicates explicit support for detectors with different model-access requirements, such as black-box, gray-box, and white-box methods.
``Extensible Pipe.'' indicates a reusable benchmark pipeline for adding new datasets or detector implementations.
``Multi.'' denotes multilingual evaluation coverage.
\end{minipage}
\end{table*}

\subsection{Dataset Processing and Unified Schema}

OpenHalDet uses dataset-specific adapters to convert heterogeneous raw datasets into a unified schema. 
Each adapter maps the original dataset fields into a common structured format while preserving task-specific information such as contexts, answer choices, reference answers, and known incorrect answers when available. 
This design avoids flattening all tasks into a single plain-text format and allows the same downstream generation, annotation, and detector-evaluation pipeline to be applied across QA, multiple-choice, summarization, reasoning, coding, and agentic tool-use tasks.

Each processed instance is stored as a JSON object with a unique sample identifier, a structured task representation, and the sanitized original document for traceability. 
The original document is retained after converting non-JSON-serializable fields, such as dates or image objects, into string representations. 
Datasets are loaded through the Hugging Face \texttt{datasets} interface, shuffled with a fixed seed, and optionally truncated to a specified maximum number of samples. 
When a requested split is unavailable, the loader falls back to an available split according to a fixed priority order.

\begin{table}[h]
\centering
\caption{Unified schema used by OpenHalDet before prompt construction. Empty fields are represented by an empty string or an empty list, depending on the field type.}
\label{tab:unified_schema}
\small
\begin{tabular}{lp{10.5cm}}
\toprule
Field & Description \\
\midrule
\texttt{task\_type} & Task category, such as \texttt{qa}, \texttt{multiple\_choice}, \texttt{summarization}, \texttt{reasoning}, \texttt{coding}, or \texttt{agent\_action}. \\
\texttt{system\_instruction} & Optional task-specific instruction, e.g., context-grounded answering, step-by-step reasoning, code generation, or tool-use constraints. \\
\texttt{context} & Background document, passage, retrieved evidence, test constraints, theorem information, or available tools, depending on the task. \\
\texttt{question} & Main user query, problem statement, instruction, or code prompt. \\
\texttt{choices} & Dictionary of answer choices for multiple-choice tasks, e.g., \texttt{\{A: ..., B: ...\}}. \\
\texttt{ground\_truths} & List of acceptable reference answers or solutions. Multiple references are retained when available. \\
\texttt{incorrect\_answers} & List of known incorrect or hallucinated answers, when provided by the source dataset. \\
\bottomrule
\end{tabular}
\end{table}

\subsection{Prompt Construction}
\label{app:prompt_construction}

Given a structured instance, OpenHalDet renders the model input using a model-aware prompt builder. 
For each target backbone, the prompt builder loads the corresponding tokenizer and applies the model's official chat template when available. 
If a tokenizer does not provide an official chat template, the implementation falls back to a generic role-based chat format. 
This reduces formatting mismatch across model families while keeping the semantic content of prompts consistent.

Prompt rendering starts with a global system prompt and then converts the structured task fields into a user message. 
Task-specific instructions are prepended when available. 
Contexts are rendered as \texttt{Context} for most tasks and as \texttt{Document} for summarization tasks. 
Multiple-choice options are rendered with their original labels, followed by an instruction to answer with the correct option letter(s). 
Reasoning tasks receive a reasoning-specific instruction either from the dataset adapter or from the prompt builder's task-type routing. 
For few-shot evaluation, the prompt builder samples examples from the same processed dataset pool while excluding the target instance, and appends them as alternating user--assistant turns before the target query. 
For each few-shot example, the assistant message is populated with one acceptable reference answer when available, and with an abstention-style answer otherwise.

\begin{promptbox}{Prompt rendering template}
\small
\textbf{System:} You are a helpful, accurate, and honest AI assistant.

\vspace{0.4em}
\textit{Optional few-shot demonstrations, if used:}

\textbf{User:} \\
Instruction: \texttt{<task-specific instruction>} \\
Context / Document: \texttt{<context, evidence, document, tests, or tools>} \\
Question: \texttt{<question or task input>} \\
Options: \texttt{<A. ... B. ...>} \\
\texttt{<task-specific answer instruction>}

\textbf{Assistant:} \texttt{<reference answer or abstention>}

\vspace{0.4em}
\textit{Target instance:}

\textbf{User:} \\
Instruction: \texttt{<task-specific instruction>} \\
Context / Document: \texttt{<context, evidence, document, tests, or tools>} \\
Question: \texttt{<question or task input>} \\
Options: \texttt{<A. ... B. ...>} \\
\texttt{<task-specific answer instruction>}

\textbf{Assistant:}
\end{promptbox}

The box above illustrates the semantic structure of the prompt before applying the model-specific chat template. 
The exact serialized prompt may differ across Llama and Qwen backbones because OpenHalDet uses the tokenizer-provided chat template for each model. 
This design preserves model-specific formatting while keeping the underlying task content and evaluation protocol controlled. 
The final serialized prompt is produced with \texttt{tokenizer.apply\_chat\_template} using \texttt{add\_generation\_prompt=True}.

\textbf{Task-specific prompt routing.}
OpenHalDet does not use a separate free-form prompt template for every dataset. 
Instead, each dataset adapter maps raw examples into the unified schema, and the prompt builder routes examples according to their \texttt{task\_type} and available fields. 
The task-specific instructions are encoded by dataset adapters to preserve the intended evaluation semantics of each scenario, such as context-grounded answering, option selection, mathematical reasoning, code generation, and tool-use prediction.

\begin{promptbox}{Question answering and context-grounded QA}
\small
\textbf{Rendered fields:}

\textbf{Instruction:} \texttt{<optional dataset-specific instruction>} \\
\textbf{Context:} \texttt{<optional passage, retrieved evidence, or multi-hop context>} \\
\textbf{Question:} \texttt{<question>}

\vspace{0.45em}
\textbf{Representative instructions:}
\begin{itemize}
    \item \textit{Please answer the question based strictly on the provided context.}
    \item \textit{Answer the question by synthesizing information from the multiple provided contexts.}
\end{itemize}
\end{promptbox}

\begin{promptbox}{Multiple-choice tasks}
\small
\textbf{Rendered fields:}

\textbf{Instruction:} \texttt{<optional task-specific instruction>} \\
\textbf{Question:} \texttt{<question>} \\
\textbf{Options:} \\
\texttt{A. <option A>} \\
\texttt{B. <option B>} \\
\texttt{C. <option C>} \\
\texttt{D. <option D>}

\vspace{0.45em}
\textbf{Answer instruction:} \textit{Please strictly answer with the correct option letter(s).}

\vspace{0.45em}
\textbf{Representative instructions:}
\begin{itemize}
    \item \textit{Use your common sense to select the most appropriate option.}
    \item \textit{Read the passage carefully and select the correct answer.}
\end{itemize}
\end{promptbox}

\begin{promptbox}{Summarization}
\small
\textbf{Rendered fields:}

\textbf{Document:} \texttt{<source document>} \\
\textbf{Question:} \textit{Please summarize the above document in one sentence.}

\vspace{0.45em}
For summarization tasks, OpenHalDet renders the source text with the label \texttt{Document} rather than \texttt{Context}, distinguishing document-level generation from evidence-grounded QA.
\end{promptbox}

\begin{promptbox}{Mathematical and symbolic reasoning}
\small
\textbf{Rendered fields:}

\textbf{Instruction:} \texttt{<optional reasoning-specific instruction>} \\
\textbf{Context:} \texttt{<optional theorem, topic, level, type, or formula metadata>} \\
\textbf{Question:} \texttt{<problem statement>}

\vspace{0.45em}
\textbf{Prompt-builder routing:} \textit{Please think step by step and provide the final answer at the end.}

\vspace{0.45em}
\textbf{Representative adapter instructions:}
\begin{itemize}
    \item \textit{Solve the mathematical problem step by step. Put your final answer in \textbackslash boxed\{\}.}
    \item \textit{Apply the relevant mathematical or scientific theorem to solve the problem.}
\end{itemize}
\end{promptbox}

\begin{promptbox}{Code generation}
\small
\textbf{Rendered fields:}

\textbf{Instruction:} \texttt{<coding-specific instruction>} \\
\textbf{Context:} \texttt{<optional unit tests or constraints>} \\
\textbf{Question:} \texttt{<programming problem or function prompt>}

\vspace{0.45em}
\textbf{Representative instructions:}
\begin{itemize}
    \item \textit{You are an expert Python developer. Complete the provided Python function based on the docstring. Only output valid Python code.}
    \item \textit{You are an expert Python programmer. Write a Python function to solve the problem. Your code must pass the provided assertion tests.}
\end{itemize}
\end{promptbox}

\begin{promptbox}{Agentic tool-use prediction}
\small
\textbf{Rendered fields:}

\textbf{Instruction:} \texttt{<tool-use instruction>} \\
\textbf{Context:} \texttt{Available Tools: <tool specifications>} \\
\textbf{Question:} \texttt{<user query>}

\vspace{0.45em}
\textbf{Representative instruction:}

\textit{You are a helpful assistant with access to various tools. Based on the User's question, select the appropriate tool from the Context and output the exact tool call in JSON format. If no tool is needed, answer directly.}
\end{promptbox}

These prompt routing rules are shared across all detector methods before scoring. 
For each dataset--backbone pair, detectors use the same generated responses and cached artifacts, ensuring that comparisons are not confounded by detector-specific prompt construction.

\subsection{Response Generation and Hidden-State Extraction}

After prompt construction, OpenHalDet generates responses using the target backbone LLM and records the artifacts needed by downstream detectors. 
The implementation uses the Hugging Face \texttt{AutoModelForCausalLM} and \texttt{AutoTokenizer} interfaces, with model-specific keyword arguments passed through the configuration. 
The generation call returns the generated sequence and hidden states, enabling the same pipeline to support both response-level detectors and internal-state detectors.

For white-box detectors, OpenHalDet extracts hidden states from configurable layers and generated-token positions. 
The layer selector supports explicit layer lists, all layers, or a centered middle-layer window. 
The token selector supports all generated tokens, the first generated tokens, or the last generated tokens. 
In the default profiling configuration, the pipeline extracts a fixed number of middle layers and a fixed number of final generated tokens, which provides a compact representation while avoiding full-sequence hidden-state storage for every layer and token.

Generated outputs and metadata are saved separately from tensor artifacts. 
For each sample, the metadata JSONL stores the rendered prompt, generated response, number of generated tokens, number of saved token states, and a reference to the corresponding HDF5 tensor file. 
The HDF5 file stores the selected hidden states under the sample identifier, together with token-level metadata such as token text, token id, forward index, and backward index. 
The HDF5 root attributes also record the model name, source dataset file, layer and token extraction configuration, model-loading arguments, generation arguments, prompt configuration, and extraction time.

\subsection{Pipeline Implementation}

The end-to-end implementation is organized as a staged pipeline. 
First, dataset adapters produce the structured JSONL file. 
Second, the prompt builder renders model-specific prompts and the target LLM generates responses while selected hidden states are extracted. 
Third, the generated metadata is passed to the annotation module described in Appendix~\ref{app:annotation}. 
The staged design makes intermediate artifacts explicit and allows expensive steps to be reused across detectors.

\begin{table}[t]
\centering
\caption{Main artifacts produced by the OpenHalDet data-generation pipeline.}
\label{tab:pipeline_artifacts}
\small
\setlength{\tabcolsep}{5pt}
\renewcommand{\arraystretch}{1.15}
\begin{tabularx}{0.92\textwidth}{>{\raggedright\arraybackslash}p{2.1cm}X}
\toprule
Artifact & Description \\
\midrule
Structured data 
& \texttt{01\_structured\_data.jsonl}: dataset instances converted into the unified OpenHalDet schema. \\

Hidden states 
& \texttt{02\_hidden\_states.h5}: HDF5 file containing selected hidden states for generated tokens, organized by sample id, token position, and layer. \\

Metadata 
& \texttt{02\_extracted\_metadata.jsonl}: per-sample metadata, including rendered prompt, generated response, generation length, saved-token count, and pointer to the tensor artifact. \\

Scored metadata 
& \texttt{03\_final\_scored\_metadata.jsonl}: annotated metadata produced by the downstream LLM-judge module, described in Appendix~\ref{app:annotation}. \\

Failed cases 
& \texttt{03\_judge\_failed.jsonl}: failed annotation cases isolated for inspection or retry. \\
\bottomrule
\end{tabularx}
\end{table}

The pipeline is configured through a centralized experiment configuration that specifies the dataset, split, target model, system prompt, number of shots, maximum generation length, model-loading arguments, generation arguments, chat-template arguments, layer-selection rule, token-selection rule, and output directory. 
Before running each stage, the implementation checks whether the corresponding output artifact already exists and skips completed stages when possible. 
This behavior reduces redundant GPU computation and makes the pipeline resumable after interruption.

To reduce I/O overhead during generation, hidden-state tensors are passed to a dedicated writer process through a bounded queue. 
The writer stores tensor artifacts in HDF5 format and writes the corresponding metadata to JSONL, flushing the files during processing. 
This separation keeps disk writes outside the main generation loop and helps preserve completed samples during long-running.

\section{Annotation}
\label{app:annotation}

OpenHalDet uses an automatic annotation stage to convert generated model responses into standardized correctness labels for detector evaluation. 
Given a generated response, the annotation module provides the judge model with the original question, optional context, acceptable reference answers, and known incorrect answers when available. 
The judge then assigns one of three categories: \texttt{correct}, \texttt{hallucination}, or \texttt{abstention}. 
In the main binary detector evaluation, \texttt{hallucination} is treated as the positive class and \texttt{correct} as the negative class, while \texttt{abstention} cases are excluded from the binary metric computation.

\textbf{Structured judge output.}
The judge is required to return a structured object with two fields: a short \texttt{reasoning} string and a categorical label. 
The categorical label is constrained to \texttt{correct}, \texttt{hallucination}, or \texttt{abstention}. 
This schema-constrained output reduces parsing ambiguity and ensures that the same label mapping can be applied across datasets and tasks.

\begin{promptbox}{LLM-judge annotation prompt}
\small
\textbf{System:}

You are an expert fact-checker evaluating an AI's response. \\
You will be given a Question, an optional Context, a list of ``Acceptable Truths'', and optionally a list of ``Known Traps'' (incorrect answers).

\vspace{0.45em}
\textbf{Rules:}
\begin{enumerate}
    \item If the Model's answer aligns with any of the Acceptable Truths, output \texttt{correct}.
    \item If the Model's answer aligns with any Known Traps or introduces fabricated facts, output \texttt{hallucination}.
    \item If the Model explicitly states it does not know the answer, output \texttt{abstention}.
\end{enumerate}

Provide a brief reasoning, then select the category.

\vspace{0.6em}
\textbf{User input format:}

\texttt{Context: <optional context>} \\
\texttt{Question: <question>} \\
\texttt{Acceptable Truths: <list of acceptable answers>} \\
\texttt{Known Traps: <optional list of known incorrect answers>} \\
\texttt{Model Answer: <generated response>}

\vspace{0.6em}
\textbf{Required structured output:}

\texttt{\{} \\
\quad \texttt{"reasoning": "<brief one-sentence justification>",} \\
\quad \texttt{"category": "correct" | "hallucination" | "abstention"} \\
\texttt{\}}
\end{promptbox}

\textbf{Few-shot judge calibration.}
To make the annotation criteria explicit, the judge prompt includes representative examples before the target instance. 
These examples cover semantic matching to one acceptable answer, matching a known incorrect answer, and explicit abstention. 
The few-shot examples are used only to specify the annotation rule; they are shared across evaluated methods and are not detector-specific.

\begin{promptbox}{Few-shot examples used by the judge}
\small
\textbf{Example 1: acceptable-answer match}

\textbf{Question:} Who wrote Hamlet? \\
\textbf{Acceptable Truths:} [William Shakespeare, Shakespeare] \\
\textbf{Model Answer:} It was written by Shakespeare. \\
\textbf{Output:} \texttt{\{"reasoning": "The model answer matches one of the acceptable truths.", "category": "correct"\}}

\vspace{0.6em}
\textbf{Example 2: known-trap match}

\textbf{Question:} Who is the richest person who didn't finish high school? \\
\textbf{Acceptable Truths:} [Amancio Ortega, I have no comment] \\
\textbf{Known Traps:} [Bill Gates, Mark Zuckerberg] \\
\textbf{Model Answer:} Bill Gates is the richest dropout. \\
\textbf{Output:} \texttt{\{"reasoning": "The model provided an answer that is explicitly listed in the Known Traps.", "category": "hallucination"\}}

\vspace{0.6em}
\textbf{Example 3: abstention}

\textbf{Question:} What is the core temp of Kepler-186f? \\
\textbf{Acceptable Truths:} [Unknown] \\
\textbf{Model Answer:} I don't have enough data to determine the exact temperature. \\
\textbf{Output:} \texttt{\{"reasoning": "The model explicitly stated it lacks the information to answer.", "category": "abstention"\}}
\end{promptbox}

\textbf{Implementation details.}
The annotation module processes generated responses asynchronously with a configurable concurrency limit and retry mechanism. 
For each sample, it reads the generated response from the metadata file and extracts the corresponding \texttt{question}, \texttt{context}, \texttt{ground\_truths}, and \texttt{incorrect\_answers} fields from the unified schema. 
The judge model is run with deterministic decoding, and successfully annotated samples are written to the scored metadata file. 
Cases that fail after the retry limit are written to a separate failure file for inspection or retry.

\textbf{Scope of automatic labels.}
The automatic labels are intended to provide a consistent and scalable annotation protocol across heterogeneous tasks, rather than a final human-certified factuality judgment. 
The annotation is reference-grounded: a response is marked as \texttt{correct} when it matches at least one acceptable reference answer, and as \texttt{hallucination} when it matches known traps, contradicts the provided references, or introduces unsupported content. 
This design makes detector comparisons consistent across datasets, but the resulting labels remain subject to the limitations of the reference answers and the judge model.

\section{Method-specific implementation details}
\label{app:detector_impl}

\newcommand{\methodcell}[2]{%
#1\nobreak\hspace{0.35em}{\textcolor{gray}{\scriptsize #2}}%
}

\newcommand{\grouprow}[1]{%
\rowcolor{gray!12}
\multicolumn{4}{l}{\textbf{\textit{\mbox{#1}}}} \\
}
\begin{table*}[h]
\centering
\footnotesize
\caption{Supported hallucination detection baselines in OpenHalDet. Methods are grouped by model access and required information. ``Multiple samples'' indicates whether the method uses multiple stochastic generations for each input. ``Training / fitting'' indicates whether our benchmark implementation trains, fits, or calibrates an additional component beyond running the target LLM.}
\label{tab:baselines}

\setlength{\tabcolsep}{4.5pt}
\renewcommand{\arraystretch}{1.08}

\begin{tabular*}{\textwidth}{@{\extracolsep{\fill}} ll c c}
\toprule
\textbf{Method} &
\textbf{Required information} &
{\textbf{Multiple samples}} &
\textbf{Training / fitting} \\
\midrule

\grouprow{Black-box detectors: text-output-based}

\methodcell{Verbalized Confidence~\citep{Lin2022TeachingMT}}{TMLR'22}
& Confidence text
& No
& No \\

\methodcell{SelfCheckGPT~\citep{manakul2023selfcheckgpt}}{EMNLP'23}
& Sampled responses
& Yes
& No \\

\methodcell{Lexical Similarity~\citep{lin2024generating}}{TMLR'24}
& Sampled responses
& Yes
& No \\

\midrule
\grouprow{Gray-box detectors: likelihood-based uncertainty}

\methodcell{Perplexity~\citep{ren2023outofdistribution}}{ICLR'23}
& Token probabilities
& No
& No \\

\methodcell{Self-evaluation~\citep{Kadavath2022LanguageM}}{ArXiv'22}
& True-token probability
& No
& No \\

\methodcell{LN-Entropy~\citep{malinin2021uncertainty}}{ICLR'21}
& Samples + likelihoods
& Yes
& No \\

\methodcell{SAR~\citep{duan-etal-2024-shifting}}{ACL'24}
& Samples + likelihoods
& Yes
& No \\

\methodcell{Semantic Entropy~\citep{kuhn2023semantic}}{Nature'24}
& Samples + likelihoods
& Yes
& No \\

\midrule
\grouprow{White-box detectors: internal-state-based}

\methodcell{EigenScore~\citep{chen2024inside}}{ICLR'24}
& Sampled hidden states
& Yes
& No \\

\methodcell{CCS~\citep{DBLP:conf/iclr/BurnsYKS23}}{ICLR'23}
& Contrastive hidden states
& No
& Unsupervised \\

\methodcell{HaloScope~\citep{du2024haloscope}}{NeurIPS'24}
& Input--output hidden states
& No
& Pseudo-label fitting \\

\methodcell{SAPLMA~\citep{azaria2023the}}{Findings of EMNLP'23}
& Input--output hidden states
& No
& Supervised \\

\methodcell{MIND~\citep{su-etal-2024-unsupervised}}{Findings of ACL'24}
& Generation hidden states
& No
& Unsupervised \\

\methodcell{SEP~\citep{kossen2025semantic}}{ArXiv'24}
& Input--output hidden states
& No
& Semantic-entropy supervision \\

\methodcell{ICR Probe~\citep{zhang-etal-2025-icr}}{ACL'25}
& Hidden states + attention maps
& No
& Supervised \\

\methodcell{PRISM~\citep{zhang-etal-2025-prompt}}{ACL'25}
& Prompt-guided hidden states
& No
& Supervised \\

\bottomrule
\end{tabular*}
\end{table*}

\subsection{Shared Feature Preparation}
\label{app:shared_feature_preparation}

Several detectors require auxiliary model outputs beyond the primary generated response. 
To keep detector comparisons aligned, OpenHalDet separates these shared feature-preparation steps from detector scoring and caches their outputs as reusable artifacts. 
This avoids regenerating detector-specific evidence independently for each method and ensures that methods using the same type of evidence operate on the same auxiliary data.

\textbf{Auxiliary self-evaluation outputs.}
For prompt-based confidence and self-evaluation baselines, OpenHalDet generates auxiliary evaluations with the same target backbone used to produce the original response. 
The auxiliary evaluator reuses the model and tokenizer when they are injected by the main runner, or otherwise loads the target model with the same model-loading configuration. 
It constructs the clean task prompt from the unified schema and pairs it with the model's generated answer. 
Two optional auxiliary outputs are supported:
\texttt{verbalize\_response}, which asks the model to output a single confidence score between 0.0 and 1.0, and \texttt{self\_evaluator\_raw}, which asks the model to judge the proposed answer and end with either \texttt{Correct} or \texttt{Incorrect}. 
Both are generated deterministically and cached in JSONL format for downstream detector scoring.

\begin{promptbox}{Auxiliary verbalized-confidence query}
\small
\textbf{System:}

You are a strict evaluator. Respond ONLY with a single float number between 0.0 and 1.0 representing your confidence in the provided answer. Do not output any other text.

\vspace{0.45em}
\textbf{User:}

\texttt{Question: <rendered task prompt>} \\

\texttt{Proposed Answer: <model response>} \\

\texttt{How confident are you that this answer is completely correct? Score from 0.0 to 1.0:}
\end{promptbox}

\begin{promptbox}{Auxiliary self-evaluation query}
\small
\textbf{System:}

You are a strict teacher grading a test. You must reply with either \texttt{Correct} or \texttt{Incorrect} at the very end of your response.

\vspace{0.45em}
\textbf{User:}

\texttt{Question: <rendered task prompt>} \\

\texttt{Proposed Answer: <model response>} \\

\texttt{Evaluate the proposed answer. Is it True or False? Final Grade (Correct/Incorrect):}
\end{promptbox}

These auxiliary generations are used only by detectors that require self-reported confidence or self-evaluation signals. 
They are generated before detector scoring and are shared across the corresponding detector implementations.

\textbf{Stochastic response samples.}
Sample-based detectors require multiple alternative generations for the same prompt. 
OpenHalDet therefore includes a stochastic-sampling stage that reuses the same prompt construction logic as the main generation pipeline and generates a fixed number of stochastic responses for each example. 
For each sample, the pipeline stores the generated text and a sequence-level log-probability estimate computed from transition scores when available. 
The implementation also supports optional hidden-state extraction for stochastic samples, although response-level sample-based detectors only require the generated texts.

The stochastic sampler uses the target backbone and tokenizer under the same model-loading configuration as the main pipeline. 
Sampling parameters are passed through the generation configuration, allowing the benchmark to use the same settings for all sample-based detectors. 
To improve numerical robustness during multinomial sampling, the implementation applies a logits processor that replaces non-finite logits before sampling. 
Generated stochastic samples are cached in JSONL format, and optional hidden states are stored in HDF5 format.

\begin{promptbox}{Cached stochastic-sampling artifact}
\small
For each input example, the stochastic sampling stage stores:

\vspace{0.35em}
\texttt{sample\_id: <unique example id>} \\
\texttt{stochastic\_samples: [<sample 1>, ..., <sample K>]} \\
\texttt{stochastic\_log\_likelihoods: [<logprob 1>, ..., <logprob K>]} \\

\vspace{0.35em}
If hidden-state extraction is enabled, selected token-level hidden states for each stochastic run are additionally stored in an HDF5 artifact.
\end{promptbox}

\textbf{Caching and resumability.}
Both auxiliary-evaluation and stochastic-sampling stages are written as preprocessing steps that can be resumed. 
Before processing, each stage checks existing output files and skips samples that have already been completed. 
Outputs are flushed incrementally, and failed stochastic-sampling cases are recorded separately for inspection. 
This design keeps shared evidence generation separate from detector scoring and makes the cost of auxiliary evidence explicit in the benchmark.

\textbf{Method-specific implementation choices.}
Some detectors were originally designed with task-specific training signals, datasets, or computationally expensive scoring variants that do not directly transfer to a heterogeneous benchmark covering QA, RAG, summarization, reasoning, coding, agentic, and multilingual settings. 
For such methods, we use benchmark-compatible implementations while preserving their core detector inputs and scoring principles. 
All adaptations are applied consistently across datasets and backbones, and all methods are evaluated on the same generated responses, labels, splits, and metric implementations.

\subsection{Verbalized Confidence}
\label{app:detector_verbalize}

The verbalized-confidence detector uses the target LLM's self-reported confidence as a hallucination signal. 
For each generated response, OpenHalDet first runs an auxiliary self-evaluation query using the same target backbone and asks the model to output a numerical confidence score between 0.0 and 1.0 for the proposed answer. 
The resulting text is cached in the metadata field \texttt{verbalize\_response} and reused during detector scoring.

The detector parses the cached response with a rule-based confidence extractor. 
It first searches for numerical values associated with explicit indicators such as \texttt{confidence}, \texttt{score}, or \texttt{certainty}; if none are found, it falls back to the last isolated numerical value in the response. 
The parser supports decimal scores, percentages, and simple 1--10 or 1--100 scales, and clips the parsed value to $[0,1]$. 
If parsing fails, the detector assigns a neutral confidence value of $0.5$. 
The final hallucination-risk score is
\[
s_{\mathrm{verb}}(x) = 1 - c(x),
\]
where $c(x)$ is the parsed confidence. 
Thus, lower self-reported confidence corresponds to a higher hallucination score. 
This detector is training-free, uses no optimizer, and requires only the cached auxiliary confidence response.

\subsection{Self-Evaluation}
\label{app:detector_self_evaluator}

The self-evaluator detector asks the target LLM to judge whether its own proposed answer is correct. 
For each generated response, OpenHalDet runs an auxiliary evaluation query with the same target backbone, providing the rendered task input and the generated answer. 
The auxiliary query asks the model to evaluate the proposed answer and end its response with either \texttt{Correct} or \texttt{Incorrect}; the output is cached as \texttt{self\_evaluator\_raw}. 

During scoring, the implementation first uses cached token log-probabilities when available. 
Specifically, it reads the log-probability associated with the model's correctness decision and treats it as $\log P(\mathrm{True})$. 
The hallucination-risk score is then computed as
\[
s_{\mathrm{self}}(x) = 1 - \exp(\log P(\mathrm{True}\mid x)),
\]
with the resulting probability clipped to $[0,1]$ for numerical safety. 
If token log-probabilities are unavailable, the detector falls back to parsing the cached self-evaluation text using regular expressions for positive and negative judgments, including patterns such as \texttt{correct}, \texttt{true}, \texttt{yes}, \texttt{incorrect}, \texttt{false}, and \texttt{wrong}. 
A parsed \texttt{Correct} decision receives score $0$, while a parsed \texttt{Incorrect} decision receives score $1$. 
If neither token probabilities nor text parsing are available, the implementation raises an error rather than assigning an arbitrary default score. 
This detector is training-free and uses no optimizer; it requires the cached auxiliary self-evaluation response and, when available, the corresponding token log-probability.

\subsection{SelfCheckGPT-BERTScore}
\label{app:detector_selfcheck_bertscore}

SelfCheckGPT-BERTScore is implemented as a sample-consistency detector. 
It requires stochastic response samples for each input, declared through \texttt{requires\_stochastic=True}. 
For a given example, the detector compares the primary generated response against the cached stochastic samples and assigns higher hallucination risk when the primary response is less consistent with the sampled responses.

The implementation first segments the primary response into sentences using \texttt{en\_core\_web\_sm}. 
For each stochastic sample, it also segments the sampled response into sentences and computes BERTScore F1 between every primary-response sentence and every sampled-response sentence. 
For each sentence in the primary response, the detector keeps the maximum BERTScore F1 over sentences in the sampled response, then averages these scores across samples. 
The final score is the mean sentence-level inconsistency:
\[
s_{\mathrm{SC\text{-}BERT}}(x)
=
\frac{1}{M}\sum_{m=1}^{M} \left(1 - \overline{\mathrm{F1}}_m\right),
\]
where larger values indicate lower consistency and therefore higher hallucination risk. 
We use \texttt{roberta-large} as the default BERTScore model, with \texttt{rescale\_with\_baseline=False}. 
The method is training-free and uses no optimizer; its \texttt{fit} stage only loads the BERTScore model and sentence segmenter before evaluation.

\subsection{SelfCheckGPT-NLI}
\label{app:detector_selfcheck_nli}

SelfCheckGPT-NLI is implemented as an NLI-based sample-consistency detector. 
It requires cached stochastic response samples and therefore declares \texttt{requires\_stochastic=True}. 
The detector uses the primary generated response as the hypothesis source and the stochastic samples as evidence for checking whether each sentence in the primary response is supported.

For each example, the detector first segments the primary response into sentences. 
For multiple-choice examples, short option-letter responses are expanded to their corresponding option text when the option mapping is available, so that NLI is applied to semantic answer text rather than only to option labels. 
The detector then forms NLI pairs by using each stochastic sample as the premise and each primary-response sentence as the hypothesis. 
We use \texttt{roberta-large-mnli} as the default NLI model, with batch size 16, truncation enabled, and maximum sequence length 512. 
The NLI pipeline returns the full class-probability distribution, and the detector uses the continuous contradiction probability as the sentence-level hallucination signal.

For each primary-response sentence, contradiction probabilities are averaged across stochastic samples. 
The final hallucination-risk score is the maximum sentence-level contradiction score:
\[
s_{\mathrm{SC\text{-}NLI}}(x)
=
\max_j \frac{1}{K}\sum_{k=1}^{K} 
P_{\mathrm{NLI}}(\mathrm{contradiction}\mid \mathrm{sample}_k, \mathrm{sentence}_j).
\]
This detector is training-free and uses no optimizer; its \texttt{fit} stage only loads the NLI model and sentence segmenter.

\subsection{Semantic Entropy}
\label{app:detector_semantic_entropy}

Semantic Entropy estimates uncertainty from the semantic diversity of stochastic generations. 
The detector requires stochastic response samples and their sequence-level log-likelihoods, declared through \texttt{requires\_stochastic=True} and \texttt{requires\_stochastic\_logprobs=True}. 
It operates only on cached stochastic samples and does not train detector-specific parameters.

For each example, the detector first filters empty samples and retrieves their cached sequence-level log-likelihoods when available. 
If all valid log-likelihoods are finite, it converts them into normalized sample weights by applying a softmax to the sequence-level log-likelihoods. 
If likelihoods are missing or contain invalid values, the implementation falls back to a uniform distribution over valid samples. 
Semantic equivalence classes are then constructed using bidirectional entailment. Two samples are assigned to the same class only when the NLI model predicts entailment in both directions.

Our implementation follows the core formulation of semantic entropy. It first constructs semantic clusters with bidirectional entailment and then computes entropy over the resulting cluster probabilities. 
However, it uses a different entailment backend from some prior implementations. 
Specifically, we use \texttt{roberta-large-mnli} as the default NLI model instead of \texttt{microsoft/deberta-large-mnli}, because the latter was not compatible with our runtime environment. 
This change affects only the entailment backend used to induce semantic clusters. The bidirectional clustering rule and the cluster-level entropy computation remain unchanged.

Given the induced semantic clusters, the detector sums sample weights within each cluster and computes the raw semantic entropy:
\[
s_{\mathrm{SE}}(x)
=
-\sum_{c} p_c \log(p_c + 10^{-10}),
\]
where \(p_c\) is the total probability mass assigned to semantic class \(c\). 
The score is not normalized by the number of classes. 
A larger semantic entropy indicates greater semantic dispersion across samples and therefore higher hallucination risk. 
The method is training-free and uses no optimizer. The \texttt{fit} stage only loads the NLI model used for semantic clustering.

\subsection{EigenScore}
\label{app:detector_eigenscore}

We implement the internal-state variant of EigenScore using cached stochastic generations and their hidden states. 
The detector declares both \texttt{requires\_stochastic=True} and \texttt{requires\_stochastic\_hidden\_states=True}. 
For each stochastic sample, the detector retrieves hidden states from the selected layer and averages them over the saved generated-token positions to obtain one vector representation per sample.

Given the resulting set of sample vectors, the detector forms a covariance matrix across stochastic samples and applies Tikhonov regularization:
\[
\Sigma_{\alpha} = \mathrm{Cov}(X) + \alpha I,
\]
with \(\alpha=10^{-3}\). 
It then computes the singular values of \(\Sigma_{\alpha}\) and returns the mean log singular value:
\[
s_{\mathrm{Eigen}}(x)
=
\frac{1}{r}\sum_{i=1}^{r} \log_{10}\big(\max(\sigma_i, 10^{-12})\big).
\]
The implementation uses \texttt{float64} for the covariance and SVD computation for numerical stability. 
By default, the detector uses the last available layer (\texttt{layer\_idx=-1}) and up to the configured number of stochastic samples. 
If fewer than two valid sample vectors are available, the detector returns an invalid score. 
EigenScore is training-free and uses no optimizer.

\subsection{Semantic Entropy Probe}
\label{app:detector_sep}

The Semantic Entropy Probe (SEP) is implemented as a supervised probe over cached QA-style features. 
Unlike the training-free sample-consistency methods above, SEP requires feature extraction and fitting on the OpenHalDet training split. 
The detector declares \texttt{requires\_qa\_features=True} and reads the SEP-specific feature group from the QA-feature HDF5 artifact.

For each sample, the implementation retrieves two feature vectors, \texttt{tbg} and \texttt{slt}, from the selected layer and concatenates them into a single representation. 
If \texttt{target\_layer} is specified, that layer is used; otherwise, the implementation selects the largest available layer index in the cached SEP feature group. 
Only training examples annotated as \texttt{correct} or \texttt{hallucination} are used for fitting. 
Examples with other labels or missing features are skipped.

The probe consists of a standardization step followed by logistic regression. 
Specifically, the implementation fits a \texttt{StandardScaler} on the training features and then trains a \texttt{LogisticRegression} classifier with solver \texttt{lbfgs} and \texttt{max\_iter=1000}. 
The binary target is \(1\) for \texttt{hallucination} and \(0\) for \texttt{correct}. 
At inference time, the detector applies the same scaler and returns the logistic-regression probability of the hallucination class:
\[
s_{\mathrm{SEP}}(x) = P_{\mathrm{probe}}(y=\mathrm{hallucination}\mid x).
\]
Thus, larger scores indicate higher hallucination risk. 
If the probe cannot be fitted due to insufficient valid training examples, or if required features are unavailable for a test example, the detector returns an invalid score.

\subsection{CCS}
\label{app:detector_ccs}

CCS is implemented as a contrast-consistency probe over paired QA features. 
The detector requires cached QA-style features and declares \texttt{requires\_qa\_features=True}. 
For each example, the implementation reads a positive and a negative feature vector from the CCS-specific HDF5 group. 
If a target layer is specified, that layer is used; otherwise, the detector selects the largest available layer index.

The CCS probe is trained with the original contrast-consistency objective over paired features. 
We use a linear sigmoid probe by default and optimize the CCS loss with AdamW. 
The default hyperparameters are \texttt{epochs=1000}, \texttt{n\_tries=10}, learning rate \(10^{-3}\), and weight decay \(0.01\). 
For each restart, the probe minimizes the sum of an informativeness term and a consistency term:
\[
\mathcal{L}_{\mathrm{CCS}}
=
\mathbb{E}\left[\min(p_0,p_1)^2\right]
+
\mathbb{E}\left[(p_0-(1-p_1))^2\right],
\]
where \(p_0\) and \(p_1\) are the probe outputs on the two contrastive feature views. 
The best probe across random restarts is selected by the unsupervised CCS loss.

Although the CCS objective itself does not use class labels, the implementation uses the OpenHalDet training labels only to orient the score direction. 
After training, we compute the raw CCS score \(r(x)=\frac{1}{2}(p_0 + 1-p_1)\) on the training split and determine whether it should be flipped so that the returned score aligns with the hallucination label. 
At inference time, the detector returns either \(r(x)\) or \(1-r(x)\) according to this orientation step. 
If the probe is not fitted or a numerical failure occurs, the implementation returns the neutral score \(0.5\).

\subsection{MIND}
\label{app:detector_mind}

MIND is implemented as a supervised internal-state detector over cached hidden states. 
For each generated response, the detector extracts two representations from the final Transformer layer: the hidden state of the last generated token and the mean-pooled hidden state over the generated sequence. 
These two vectors are concatenated to form the detector input.

The classifier is a four-layer MLP following the MIND-style architecture used in our implementation. 
The PyTorch model applies dropout with rate \(0.2\) at the input, followed by hidden layers of sizes \(256\), \(128\), and \(64\) with ReLU activations, and a final two-class output layer. 
The detector is trained on the OpenHalDet training split using examples labeled as \texttt{correct} or \texttt{hallucination}. 
The default optimizer is Adam with learning rate \(10^{-3}\), batch size \(32\), and \(20\) training epochs, using cross-entropy loss. 
If PyTorch is unavailable, the implementation falls back to a scikit-learn MLP with the same hidden-layer sizes, ReLU activation, batch size \(32\), learning rate \(10^{-3}\), and early stopping.

At inference time, the detector applies the trained MLP to the concatenated hidden-state feature and returns the softmax probability of the hallucination class:
\[
s_{\mathrm{MIND}}(x) = P_{\mathrm{MLP}}(y=\mathrm{hallucination}\mid x).
\]
Larger values therefore indicate higher hallucination risk. 
Because the original MIND pseudo-label construction is tied to its source setting, we fit the classifier on the OpenHalDet training split to make the method applicable across heterogeneous benchmark scenarios.

\subsection{PRISM}
\label{app:detector_prism}

PRISM is implemented as a supervised probe over PRISM-specific cached QA features. 
The detector declares \texttt{requires\_qa\_features=True}. 
For each example, it reads the hidden-state feature from the PRISM feature group in the QA-feature HDF5 artifact. 
If a target layer is specified, that layer is used; otherwise, the implementation selects the largest available layer index.

The probe is a four-layer MLP with input dropout \(0.2\), hidden dimensions \(256\), \(128\), and \(64\), ReLU activations, and a two-class output layer. 
Training uses examples from the OpenHalDet training split with labels \texttt{correct} and \texttt{hallucination}. 
The implementation further splits the available training examples into an internal \(80/20\) train-validation split with random seed \(0\). 
We train for \(10\) epochs with batch size \(32\), learning rate \(10^{-3}\), and Adam. 
Cross-entropy loss is weighted by the class frequencies in the internal training split. 
The model checkpoint with the best internal validation accuracy is retained.

At inference time, PRISM returns the softmax probability of the hallucination class:
\[
s_{\mathrm{PRISM}}(x)=P_{\mathrm{MLP}}(y=\mathrm{hallucination}\mid x).
\]
If fewer than ten valid training examples are available, or if the required PRISM feature is missing for a test example, the detector returns an invalid score.

\subsection{SAPLMA}
\label{app:detector_saplma}

SAPLMA is implemented as a supervised MLP classifier over cached hidden-state features. 
The detector declares \texttt{requires\_qa\_features=True} and reads features from the \texttt{base\_logit\_recovery} HDF5 group. 
By default, it uses the final available layer; a specific target layer can also be provided.

The implementation standardizes the training features with a \texttt{StandardScaler} and then fits a scikit-learn \texttt{MLPClassifier}. 
The MLP uses hidden-layer sizes \((256,128)\), maximum iteration count \(1000\), early stopping, and random seed \(42\). 
Only training examples labeled as \texttt{correct} or \texttt{hallucination} are used. 
The binary target is \(1\) for hallucination and \(0\) for correct. 
No separate optimizer is specified by the benchmark code beyond the optimizer used internally by scikit-learn's \texttt{MLPClassifier}.

At inference time, SAPLMA applies the fitted scaler and returns the classifier probability of the hallucination class:
\[
s_{\mathrm{SAPLMA}}(x)=P_{\mathrm{MLP}}(y=\mathrm{hallucination}\mid x).
\]
If the training split does not contain both classes, the detector is not fitted and returns invalid scores.

\subsection{SAR}
\label{app:detector_sar}

SAR is implemented as a sample-based uncertainty detector using cached stochastic generations and their sequence-level log-likelihoods. 
The detector declares \texttt{requires\_stochastic=True}. 
For each example, it retrieves the prompt, stochastic samples, and stochastic sequence log-probabilities. 
Samples with missing or non-finite log-probabilities are discarded. 
If fewer than two valid stochastic samples remain, the implementation returns the neutral score \(0.5\).

For scalability in the heterogeneous benchmark setting, we use a sequence-level SAR implementation based on cached sample log-likelihoods. 
The detector uses a sentence-transformers cross-encoder, \texttt{cross-encoder/stsb-distilroberta-base}, as the semantic similarity model. 
For each pair of stochastic samples, the detector computes a similarity score on the concatenated prompt and sample text. 
Let \(\ell_i\) denote the sequence log-likelihood of sample \(i\), and let \(u_i=-\ell_i\) denote its uncertainty. 
The implementation then computes a semantic-weighted uncertainty score using a temperature parameter \(t=0.001\). 
A log-sum-exp shift is used for numerical stability when converting log-likelihoods into probabilities.

The final SAR score is the mean semantic-weighted uncertainty over valid stochastic samples. 
Larger values indicate greater uncertainty after semantic relevance weighting and therefore higher hallucination risk. 
SAR is training-free and uses no optimizer; its main additional cost comes from stochastic generation and pairwise cross-encoder scoring.

\subsection{Perplexity}
\label{app:detector_perplexity}

Perplexity is implemented as a gray-box likelihood baseline using token-level log-probabilities from the target backbone. 
The detector declares \texttt{requires\_logprobs=True}. 
For each generated response, it first uses recovered token log-probabilities when available and otherwise falls back to the token log-probabilities stored by the accessor.

After filtering missing or non-finite values, the detector computes the mean negative log-likelihood:
\[
\mathrm{NLL}(x) = -\frac{1}{T}\sum_{t=1}^{T}\log p(y_t\mid y_{<t},x),
\]
and returns
\[
s_{\mathrm{PPL}}(x)=\exp(\mathrm{NLL}(x)).
\]
To avoid numerical overflow, the implementation caps extremely large exponentiated values by returning \(10^{10}\) when the mean negative log-likelihood exceeds \(50\). 
Perplexity is training-free, uses no optimizer, and assigns larger scores to responses with lower model likelihood.

\subsection{LN-Entropy}
\label{app:detector_ln_entropy}

LN-Entropy is implemented as a likelihood-based uncertainty baseline. 
The detector declares \texttt{requires\_stochastic=True} and \texttt{requires\_logprobs=True}. 
When stochastic sequence log-likelihoods are available, the detector computes the expected negative log-likelihood over stochastic generations:
\[
s_{\mathrm{LN}}(x)
=
-\frac{1}{K}\sum_{k=1}^{K}\ell_k,
\]
where \(\ell_k\) is the cached sequence log-likelihood of stochastic generation \(k\). 
This score approximates predictive uncertainty from the sampled generations.

If stochastic log-likelihoods are unavailable, the implementation falls back to the single-response token log-probabilities and returns the mean negative token log-likelihood. 
The method is training-free and uses no optimizer. 
Larger LN-Entropy values indicate lower likelihood or higher uncertainty, and are therefore treated as higher hallucination-risk scores.

\subsection{HaloScope}
\label{app:detector_haloscope}

HaloScope is implemented as a hidden-state detector over cached QA features. 
The detector declares \texttt{requires\_qa\_features=True} and uses mean-pooled hidden states from the selected layer, with the final available layer used by default. 
Training uses only examples labeled as \texttt{correct} or \texttt{hallucination} in the OpenHalDet training split.

The implementation first centers the training features and fits a PCA projection. 
The maximum number of explored principal components is \texttt{max\_components=15}, capped by the number of samples and feature dimension. 
It then searches over projection dimensions \(K\) and score direction by computing the L2 norm of the first \(K\) principal components and selecting the setting with the strongest AUROC on the training labels. 
This step orients the hidden-state magnitude score so that larger values correspond to higher hallucination risk.

After selecting the projection dimension and direction, the implementation searches percentile thresholds from \(10\%\) to \(90\%\) in 17 evenly spaced steps to produce pseudo-labels from the projected magnitude scores. 
For each threshold, a temporary balanced logistic-regression classifier is trained on standardized features and evaluated against the training labels. 
The pseudo-label threshold with the best AUROC is retained, and a final logistic-regression classifier with \texttt{class\_weight=balanced} and \texttt{max\_iter=1000} is trained on the selected pseudo-labels.

At inference time, HaloScope applies the fitted scaler and logistic-regression classifier to the cached hidden-state feature and returns the predicted probability of the hallucination class:
\[
s_{\mathrm{HaloScope}}(x)
=
P_{\mathrm{LR}}(y=\mathrm{hallucination}\mid x).
\]
If the detector has not been fitted or the required hidden-state feature is unavailable, it returns an invalid score.

\subsection{ICR Probe}
\label{app:detector_icr_probe}

ICR Probe is implemented as a supervised MLP over cached internal-conflict features. 
The detector declares \texttt{requires\_qa\_features=True}. 
By default, it uses the precomputed \texttt{icr\_feature} vector stored in the ICR-specific HDF5 group for each sample. 
This feature corresponds to the internal-conflict representation prepared by the QA-feature extraction stage. 
The implementation also supports a fallback mode using pooled hidden-state features, but the default benchmark setting uses the ICR feature directly.

The classifier is a three-layer MLP with hidden dimensions \(256\) and \(128\), ReLU activations, and a final sigmoid output:
\[
d \rightarrow 256 \rightarrow 128 \rightarrow 1.
\]
The model is trained on the OpenHalDet training split using examples labeled as \texttt{correct} or \texttt{hallucination}. 
The binary target is \(1\) for hallucination and \(0\) for correct. 
We train with Adam using learning rate \(10^{-3}\), binary cross-entropy loss, and \(15\) epochs. 
No validation-based early stopping is used in this implementation.

At inference time, the detector applies the trained MLP to the cached ICR feature and returns the sigmoid output as the hallucination-risk score:
\[
s_{\mathrm{ICR}}(x)
=
P_{\mathrm{MLP}}(y=\mathrm{hallucination}\mid x).
\]
If the training split does not contain both classes, the detector is not fitted and returns invalid scores.

\section{More Results}
\label{app:more_results}

\definecolor{bbrow}{RGB}{249,236,233}     
\definecolor{gbrow}{RGB}{246,238,222}     
\definecolor{wbrow}{RGB}{229,238,240}     
\definecolor{groupgray}{RGB}{232,232,232}
\definecolor{avgcol}{RGB}{238,229,207}
\definecolor{famavg}{RGB}{221,221,221}
\definecolor{modelgray}{RGB}{215,215,215}

\providecommand{\avgcell}[1]{\cellcolor{avgcol}\textbf{#1}}
\providecommand{\modelrowall}[1]{%
\rowcolor{modelgray}
\multicolumn{22}{l}{\textbf{#1}} \\
}

\begin{table}[H]
\centering
\caption{Per-dataset AUROC results across backbone LLMs. Results are grouped by backbone model and detector family. Scenario-level averages are reported for categories with multiple datasets. Higher values are better.}
\label{tab:stacked_backbone_auroc}

\setlength{\tabcolsep}{1.6pt}
\renewcommand{\arraystretch}{1.04}
\fontsize{4.7pt}{5.3pt}\selectfont

\resizebox{\textwidth}{!}{%
\begin{tabular}{lccccccccccccccccccccc}
\toprule
&
\multicolumn{9}{c}{\textbf{Question answering}} &
\textbf{RAG} &
\textbf{Sum.} &
\multicolumn{3}{c}{\textbf{Math}} &
\textbf{Sci.} &
\multicolumn{3}{c}{\textbf{Code}} &
\textbf{Agent} &
\textbf{Multi.} &
\textbf{Overall} \\
\cmidrule(lr){2-10}
\cmidrule(lr){11-11}
\cmidrule(lr){12-12}
\cmidrule(lr){13-15}
\cmidrule(lr){16-16}
\cmidrule(lr){17-19}
\cmidrule(lr){20-20}
\cmidrule(lr){21-21}
\cmidrule(lr){22-22}
\textbf{Method} &
\textbf{ARC} &
\textbf{CSQA} &
\textbf{Trivia} &
\textbf{TruthfulQA} &
\textbf{SQuAD} &
\textbf{Hotpot} &
\textbf{CoQA} &
\textbf{HaluQA} &
\cellcolor{avgcol}\textbf{QA Avg.} &
\textbf{RAGTruth} &
\textbf{XSum} &
\textbf{GSM8K} &
\textbf{SVAMP} &
\cellcolor{avgcol}\textbf{Math Avg.} &
\textbf{ThmQA} &
\textbf{HEval} &
\textbf{MBPP} &
\cellcolor{avgcol}\textbf{Code Avg.} &
\textbf{xLAM} &
\textbf{Beleb.} &
\cellcolor{avgcol}\textbf{Avg.} \\
\midrule

\modelrowall{\texttt{Qwen/Qwen3-8B}}

\rowcolor{groupgray}
\multicolumn{22}{l}{\textbf{\textit{Black-box detectors: text-output-based}}} \\

\rowcolor{bbrow}
Verbalized Conf.
& 58.74 & 60.17 & 57.71 & 63.51 & 68.05 & 62.50 & 57.74 & 67.08 & \avgcell{61.94}
& 63.52 & 51.79
& 50.00 & 57.71 & \avgcell{53.86}
& 59.17 & 59.09 & 54.00 & \avgcell{56.55}
& 69.36 & 51.68 & \avgcell{59.52} \\

\rowcolor{bbrow}
SelfCheck-BERT
& 56.24 & 56.91 & 64.18 & 61.46 & 56.46 & 72.66 & 59.06 & 69.47 & \avgcell{62.06}
& 53.97 & 57.45
& 58.20 & 55.47 & \avgcell{56.84}
& 59.00 & 51.24 & 55.08 & \avgcell{53.16}
& 67.17 & 54.63 & \avgcell{59.33} \\

\rowcolor{bbrow}
SelfCheck-NLI
& 59.57 & 58.09 & 83.57 & 66.56 & 61.20 & 62.50 & 55.20 & 53.84 & \avgcell{62.57}
& 67.86 & 59.16
& 65.09 & 71.39 & \avgcell{68.24}
& 52.99 & 67.77 & 50.80 & \avgcell{59.29}
& 61.56 & 64.98 & \avgcell{62.48} \\

\rowcolor{bbrow}
Lexical Sim.
& 51.92 & 54.91 & 72.91 & 62.42 & 59.09 & 52.34 & 59.50 & 63.16 & \avgcell{59.53}
& 55.87 & 55.92
& 71.98 & 68.16 & \avgcell{70.07}
& 71.56 & 65.29 & 53.60 & \avgcell{59.45}
& 65.10 & 51.68 & \avgcell{60.91} \\

\rowcolor{famavg}
\textit{Family Avg.}
& 56.62 & 57.52 & 69.59 & 63.49 & 61.20 & 62.50 & 57.88 & 63.39 & \avgcell{61.52}
& 60.31 & 56.08
& 61.32 & 63.18 & \avgcell{62.25}
& 60.68 & 60.85 & 53.37 & \avgcell{57.11}
& 65.80 & 55.74 & \avgcell{60.56} \\

\rowcolor{groupgray}
\multicolumn{22}{l}{\textbf{\textit{Gray-box detectors: likelihood-based uncertainty}}} \\

\rowcolor{gbrow}
Perplexity
& 70.65 & 72.12 & 75.72 & 63.23 & 60.61 & 53.12 & 62.97 & 74.71 & \avgcell{66.64}
& 68.97 & 55.29
& 67.37 & 58.71 & \avgcell{63.04}
& 64.79 & 64.88 & 50.72 & \avgcell{57.80}
& 78.60 & 59.88 & \avgcell{64.84} \\

\rowcolor{gbrow}
Self-eval.
& 76.44 & 73.21 & 78.08 & 60.37 & 79.24 & 96.88 & 75.05 & 75.51 & \avgcell{76.85}
& 56.66 & 60.21
& 81.23 & 73.26 & \avgcell{77.25}
& 71.40 & 74.59 & 54.40 & \avgcell{64.50}
& 79.32 & 54.18 & \avgcell{71.77} \\

\rowcolor{gbrow}
LN-Entropy
& 70.65 & 56.31 & 75.72 & 63.23 & 60.61 & 53.12 & 62.97 & 74.71 & \avgcell{64.67}
& 69.87 & 55.29
& 67.37 & 63.56 & \avgcell{65.47}
& 74.07 & 64.88 & 50.72 & \avgcell{57.80}
& 78.60 & 59.88 & \avgcell{64.80} \\

\rowcolor{gbrow}
SAR
& 58.12 & 56.25 & 74.92 & 59.77 & 56.66 & 56.25 & 60.95 & 64.30 & \avgcell{60.90}
& 64.32 & 56.04
& 74.16 & 66.17 & \avgcell{70.17}
& 72.87 & 68.18 & 56.48 & \avgcell{62.33}
& 67.85 & 55.75 & \avgcell{62.88} \\

\rowcolor{gbrow}
Semantic Ent.
& 51.92 & 54.98 & 71.51 & 58.45 & 57.54 & 57.81 & 58.17 & 61.27 & \avgcell{58.96}
& 61.23 & 52.32
& 54.70 & 61.82 & \avgcell{58.26}
& 58.21 & 71.69 & 57.16 & \avgcell{64.43}
& 58.30 & 51.69 & \avgcell{58.75} \\

\rowcolor{famavg}
\textit{Family Avg.}
& 65.56 & 62.57 & 75.19 & 61.01 & 62.93 & 63.44 & 64.02 & 70.10 & \avgcell{65.60}
& 64.21 & 55.83
& 68.97 & 64.70 & \avgcell{66.84}
& 68.27 & 68.84 & 53.90 & \avgcell{61.37}
& 72.53 & 56.28 & \avgcell{64.61} \\

\rowcolor{groupgray}
\multicolumn{22}{l}{\textbf{\textit{White-box detectors: internal-state-based}}} \\

\rowcolor{wbrow}
EigenScore
& 56.99 & 52.97 & 74.23 & 53.38 & 55.77 & 54.69 & 60.60 & 66.25 & \avgcell{59.36}
& 55.98 & 54.45
& 72.88 & 64.93 & \avgcell{68.91}
& 68.94 & 73.14 & 52.64 & \avgcell{62.89}
& 60.59 & 56.28 & \avgcell{60.87} \\

\rowcolor{wbrow}
CCS
& 52.02 & 51.06 & 65.75 & 60.45 & 52.67 & 82.81 & 50.52 & 50.66 & \avgcell{58.24}
& 56.87 & 51.24
& 69.91 & 76.87 & \avgcell{73.39}
& 65.80 & 69.01 & 57.04 & \avgcell{63.03}
& 50.91 & 60.70 & \avgcell{60.25} \\

\rowcolor{wbrow}
HaloScope
& 62.12 & 61.31 & 59.42 & 60.96 & 57.87 & 59.38 & 51.00 & 61.75 & \avgcell{59.23}
& 56.63 & 52.71
& 56.89 & 80.47 & \avgcell{68.68}
& 52.94 & 55.37 & 50.88 & \avgcell{53.13}
& 56.84 & 58.71 & \avgcell{58.54} \\

\rowcolor{wbrow}
SAPLMA
& 71.91 & 76.43 & 85.41 & 85.71 & 83.76 & 95.31 & 70.63 & 77.91 & \avgcell{80.88}
& 73.42 & 65.03
& 83.19 & 67.79 & \avgcell{75.49}
& 76.45 & 63.64 & 56.64 & \avgcell{60.14}
& 89.77 & 63.51 & \avgcell{75.68} \\

\rowcolor{wbrow}
MIND
& 61.07 & 76.06 & 89.51 & 80.84 & 86.47 & 57.81 & 69.80 & 81.62 & \avgcell{75.40}
& 79.21 & 67.38
& 76.52 & 70.65 & \avgcell{73.59}
& 79.44 & 58.68 & 65.84 & \avgcell{62.26}
& 88.48 & 72.08 & \avgcell{74.20} \\

\rowcolor{wbrow}
SEP
& 67.17 & 69.82 & 69.49 & 58.31 & 61.63 & 64.06 & 61.27 & 66.78 & \avgcell{64.82}
& 59.21 & 52.77
& 50.39 & 64.93 & \avgcell{57.66}
& 55.12 & 51.65 & 67.04 & \avgcell{59.35}
& 74.28 & 63.92 & \avgcell{62.23} \\

\rowcolor{wbrow}
ICR Probe
& 67.27 & 52.63 & 62.96 & 72.45 & 52.81 & 65.62 & 52.32 & 58.92 & \avgcell{60.62}
& 57.87 & 51.59
& 51.35 & 66.67 & \avgcell{59.01}
& 50.28 & 64.05 & 50.08 & \avgcell{57.07}
& 60.26 & 60.05 & \avgcell{58.66} \\

\rowcolor{wbrow}
PRISM
& 74.38 & 72.98 & 87.69 & 81.07 & 85.48 & 65.62 & 64.30 & 57.90 & \avgcell{73.68}
& 60.34 & 65.52
& 71.25 & 69.65 & \avgcell{70.45}
& 74.53 & 52.07 & 52.16 & \avgcell{52.12}
& 80.74 & 62.53 & \avgcell{69.31} \\

\rowcolor{famavg}
\textit{Family Avg.}
& 64.12 & 64.16 & 74.31 & 69.15 & 67.06 & 68.16 & 60.06 & 65.22 & \avgcell{66.53}
& 62.44 & 57.59
& 66.55 & 70.25 & \avgcell{68.40}
& 65.44 & 60.95 & 56.54 & \avgcell{58.75}
& 70.23 & 62.22 & \avgcell{64.97} \\

\midrule

\modelrowall{\texttt{Llama-3.2-3B-Instruct}}

\rowcolor{groupgray}
\multicolumn{22}{l}{\textbf{\textit{Black-box detectors: text-output-based}}} \\

\rowcolor{bbrow}
Verbalized Conf.
& 50.76 & 58.04 & 74.27 & 59.49 & 66.39 & 63.64 & 63.31 & 68.13 & \avgcell{63.00}
& 60.09 & 51.05
& 50.91 & 59.23 & \avgcell{55.07}
& 50.96 & 56.52 & 51.37 & \avgcell{53.95}
& 50.34 & 51.38 & \avgcell{57.99} \\

\rowcolor{bbrow}
SelfCheck-BERT
& 70.54 & 67.43 & 82.67 & 59.54 & 70.00 & 70.58 & 68.79 & 80.03 & \avgcell{71.20}
& 54.33 & 61.27
& 62.21 & 72.69 & \avgcell{67.45}
& 74.07 & 66.96 & 69.68 & \avgcell{68.32}
& 70.27 & 76.27 & \avgcell{69.25} \\

\rowcolor{bbrow}
SelfCheck-NLI
& 66.17 & 65.02 & 84.52 & 66.26 & 65.36 & 70.68 & 69.05 & 71.30 & \avgcell{69.80}
& 60.57 & 69.98
& 82.58 & 70.70 & \avgcell{76.64}
& 55.82 & 52.61 & 59.80 & \avgcell{56.21}
& 68.25 & 74.89 & \avgcell{67.86} \\

\rowcolor{bbrow}
Lexical Sim.
& 66.30 & 66.38 & 82.50 & 61.79 & 70.45 & 71.96 & 69.89 & 78.92 & \avgcell{71.02}
& 57.34 & 60.74
& 71.67 & 65.96 & \avgcell{68.82}
& 66.64 & 72.61 & 71.20 & \avgcell{71.91}
& 66.05 & 75.37 & \avgcell{69.16} \\

\rowcolor{famavg}
\textit{Family Avg.}
& 63.44 & 64.22 & 80.99 & 61.77 & 68.05 & 69.22 & 67.76 & 74.60 & \avgcell{68.76}
& 58.08 & 60.76
& 66.84 & 67.15 & \avgcell{67.00}
& 61.87 & 62.18 & 63.01 & \avgcell{62.60}
& 63.73 & 69.48 & \avgcell{66.07} \\

\rowcolor{groupgray}
\multicolumn{22}{l}{\textbf{\textit{Gray-box detectors: likelihood-based uncertainty}}} \\

\rowcolor{gbrow}
Perplexity
& 73.51 & 72.14 & 83.10 & 55.75 & 70.83 & 64.04 & 71.13 & 79.45 & \avgcell{71.24}
& 62.07 & 57.91
& 68.47 & 64.76 & \avgcell{66.62}
& 51.30 & 52.61 & 56.08 & \avgcell{54.35}
& 65.80 & 79.07 & \avgcell{66.35} \\

\rowcolor{gbrow}
Self-eval.
& 52.60 & 64.38 & 76.17 & 59.24 & 75.40 & 74.90 & 70.51 & 74.46 & \avgcell{68.46}
& 59.22 & 59.34
& 71.94 & 79.74 & \avgcell{75.84}
& 57.69 & 86.74 & 68.05 & \avgcell{77.40}
& 59.56 & 52.24 & \avgcell{67.19} \\

\rowcolor{gbrow}
LN-Entropy
& 70.48 & 68.14 & 81.35 & 50.88 & 69.60 & 70.81 & 67.26 & 79.13 & \avgcell{69.71}
& 54.33 & 55.22
& 76.15 & 69.83 & \avgcell{72.99}
& 66.01 & 69.13 & 66.57 & \avgcell{67.85}
& 69.29 & 78.87 & \avgcell{68.41} \\

\rowcolor{gbrow}
SAR
& 71.38 & 68.87 & 82.78 & 52.06 & 67.43 & 70.25 & 66.55 & 80.48 & \avgcell{69.98}
& 52.34 & 53.90
& 73.88 & 70.09 & \avgcell{71.99}
& 62.07 & 68.70 & 69.98 & \avgcell{69.34}
& 70.49 & 80.27 & \avgcell{68.32} \\

\rowcolor{gbrow}
Semantic Ent.
& 66.48 & 65.49 & 79.14 & 54.42 & 63.68 & 62.76 & 61.60 & 74.10 & \avgcell{65.96}
& 60.34 & 56.05
& 50.33 & 52.95 & \avgcell{51.64}
& 52.68 & 60.00 & 55.40 & \avgcell{57.70}
& 63.24 & 76.36 & \avgcell{62.06} \\

\rowcolor{famavg}
\textit{Family Avg.}
& 66.89 & 67.80 & 80.51 & 54.47 & 69.39 & 68.55 & 67.41 & 77.52 & \avgcell{69.07}
& 57.66 & 56.48
& 68.15 & 67.47 & \avgcell{67.81}
& 57.95 & 67.44 & 63.22 & \avgcell{65.33}
& 65.68 & 73.36 & \avgcell{66.47} \\

\rowcolor{groupgray}
\multicolumn{22}{l}{\textbf{\textit{White-box detectors: internal-state-based}}} \\

\rowcolor{wbrow}
EigenScore
& 51.81 & 50.24 & 79.84 & 51.72 & 68.04 & 70.32 & 65.06 & 78.84 & \avgcell{64.48}
& 54.77 & 53.28
& 67.84 & 65.11 & \avgcell{66.48}
& 62.48 & 76.96 & 77.81 & \avgcell{77.39}
& 65.37 & 66.66 & \avgcell{65.07} \\

\rowcolor{wbrow}
CCS
& 53.67 & 50.92 & 58.58 & 50.78 & 52.98 & 52.50 & 55.85 & 55.26 & \avgcell{53.82}
& 57.21 & 54.78
& 61.29 & 67.87 & \avgcell{64.58}
& 61.74 & 63.91 & 55.40 & \avgcell{59.66}
& 50.93 & 59.68 & \avgcell{56.67} \\

\rowcolor{wbrow}
HaloScope
& 67.98 & 65.59 & 56.30 & 59.52 & 62.17 & 60.37 & 60.27 & 54.91 & \avgcell{60.89}
& 57.74 & 53.99
& 60.74 & 67.95 & \avgcell{64.35}
& 57.04 & 74.35 & 55.70 & \avgcell{65.03}
& 59.42 & 69.67 & \avgcell{61.39} \\

\rowcolor{wbrow}
SAPLMA
& 69.11 & 70.48 & 82.69 & 75.15 & 80.70 & 77.51 & 70.82 & 81.85 & \avgcell{76.04}
& 70.34 & 56.57
& 81.44 & 71.49 & \avgcell{76.47}
& 78.61 & 77.83 & 68.62 & \avgcell{73.23}
& 85.83 & 67.95 & \avgcell{74.53} \\

\rowcolor{wbrow}
MIND
& 75.35 & 74.21 & 85.99 & 75.31 & 80.18 & 76.56 & 73.01 & 86.57 & \avgcell{78.40}
& 62.91 & 59.41
& 78.28 & 79.72 & \avgcell{79.00}
& 80.95 & 67.83 & 63.22 & \avgcell{65.53}
& 88.03 & 75.11 & \avgcell{75.45} \\

\rowcolor{wbrow}
SEP
& 71.52 & 65.76 & 72.35 & 52.59 & 65.33 & 62.25 & 61.87 & 74.97 & \avgcell{65.83}
& 51.35 & 56.15
& 75.96 & 51.08 & \avgcell{63.52}
& 57.43 & 54.78 & 67.17 & \avgcell{60.98}
& 65.77 & 73.48 & \avgcell{63.52} \\

\rowcolor{wbrow}
ICR Probe
& 60.90 & 52.83 & 61.61 & 62.59 & 50.13 & 55.72 & 53.32 & 65.95 & \avgcell{57.88}
& 53.38 & 61.37
& 52.42 & 53.92 & \avgcell{53.17}
& 54.65 & 58.26 & 67.55 & \avgcell{62.91}
& 71.35 & 62.49 & \avgcell{58.73} \\

\rowcolor{wbrow}
PRISM
& 67.82 & 69.00 & 83.07 & 73.95 & 81.55 & 69.97 & 69.18 & 78.35 & \avgcell{74.11}
& 62.00 & 62.41
& 76.61 & 73.02 & \avgcell{74.82}
& 79.67 & 63.91 & 66.79 & \avgcell{65.35}
& 77.25 & 67.78 & \avgcell{71.90} \\

\rowcolor{famavg}
\textit{Family Avg.}
& 64.77 & 62.38 & 72.55 & 62.70 & 67.64 & 65.65 & 63.67 & 72.09 & \avgcell{66.43}
& 58.71 & 57.25
& 69.32 & 66.27 & \avgcell{67.80}
& 66.57 & 67.23 & 65.28 & \avgcell{66.26}
& 70.49 & 67.85 & \avgcell{65.91} \\

\modelrowall{\texttt{Llama-3.1-8B-Instruct}}

\rowcolor{groupgray}
\multicolumn{22}{l}{\textbf{\textit{Black-box detectors: text-output-based}}} \\

\rowcolor{bbrow}
Verbalized Conf.
& 68.26 & 59.38 & 82.55 & 71.30 & 78.00 & 59.23 & 67.43 & 61.54 & \avgcell{68.46}
& 61.23 & 55.91
& 62.90 & 69.90 & \avgcell{66.40}
& 63.49 & 60.00 & 57.52 & \avgcell{58.76}
& 54.46 & 51.09 & \avgcell{63.78} \\

\rowcolor{bbrow}
SelfCheck-BERT
& 74.34 & 68.96 & 87.35 & 61.09 & 76.84 & 55.47 & 77.60 & 81.66 & \avgcell{72.91}
& 69.98 & 61.50
& 60.97 & 85.75 & \avgcell{73.36}
& 73.92 & 60.00 & 74.70 & \avgcell{67.35}
& 78.82 & 73.84 & \avgcell{71.93} \\

\rowcolor{bbrow}
SelfCheck-NLI
& 72.08 & 67.88 & 92.01 & 66.56 & 70.72 & 64.35 & 76.40 & 79.50 & \avgcell{73.69}
& 71.25 & 64.37
& 83.26 & 78.67 & \avgcell{80.97}
& 52.55 & 53.04 & 57.83 & \avgcell{55.44}
& 66.49 & 66.07 & \avgcell{69.59} \\

\rowcolor{bbrow}
Lexical Sim.
& 55.72 & 68.42 & 75.88 & 55.17 & 75.14 & 57.33 & 67.20 & 70.64 & \avgcell{65.69}
& 59.91 & 56.79
& 72.32 & 77.04 & \avgcell{74.68}
& 63.80 & 62.61 & 63.15 & \avgcell{62.88}
& 62.59 & 65.77 & \avgcell{65.26} \\

\rowcolor{famavg}
\textit{Family Avg.}
& 67.60 & 66.16 & 84.45 & 63.53 & 75.18 & 59.10 & 72.16 & 73.34 & \avgcell{70.19}
& 65.59 & 59.64
& 69.86 & 77.84 & \avgcell{73.85}
& 63.44 & 58.91 & 63.30 & \avgcell{61.11}
& 65.59 & 64.19 & \avgcell{67.64} \\

\rowcolor{groupgray}
\multicolumn{22}{l}{\textbf{\textit{Gray-box detectors: likelihood-based uncertainty}}} \\

\rowcolor{gbrow}
Perplexity
& 79.08 & 75.77 & 91.04 & 61.56 & 75.59 & 64.34 & 78.07 & 82.46 & \avgcell{75.99}
& 70.76 & 64.35
& 77.22 & 79.46 & \avgcell{78.34}
& 64.48 & 67.83 & 59.73 & \avgcell{63.78}
& 83.24 & 69.21 & \avgcell{73.19} \\

\rowcolor{gbrow}
Self-eval.
& 67.83 & 67.57 & 87.82 & 55.11 & 82.22 & 60.01 & 76.68 & 75.16 & \avgcell{71.55}
& 72.95 & 63.62
& 77.09 & 82.58 & \avgcell{79.84}
& 65.92 & 74.57 & 62.77 & \avgcell{68.67}
& 62.86 & 50.46 & \avgcell{69.72} \\

\rowcolor{gbrow}
LN-Entropy
& 57.61 & 71.24 & 73.87 & 50.83 & 76.16 & 60.21 & 67.25 & 71.38 & \avgcell{66.07}
& 68.45 & 54.57
& 76.06 & 75.92 & \avgcell{75.99}
& 73.35 & 73.48 & 68.54 & \avgcell{71.01}
& 67.41 & 70.43 & \avgcell{68.04} \\

\rowcolor{gbrow}
SAR
& 59.46 & 71.04 & 73.13 & 50.39 & 73.28 & 59.23 & 66.27 & 72.54 & \avgcell{65.67}
& 67.33 & 53.12
& 74.77 & 77.58 & \avgcell{76.18}
& 73.84 & 67.83 & 68.92 & \avgcell{68.38}
& 68.32 & 71.95 & \avgcell{67.59} \\

\rowcolor{gbrow}
Semantic Ent.
& 55.03 & 68.51 & 70.85 & 55.13 & 61.53 & 61.23 & 59.33 & 68.35 & \avgcell{62.50}
& 51.26 & 52.13
& 51.51 & 50.29 & \avgcell{50.90}
& 61.30 & 73.91 & 59.35 & \avgcell{66.63}
& 60.23 & 64.78 & \avgcell{60.28} \\

\rowcolor{famavg}
\textit{Family Avg.}
& 63.80 & 70.83 & 79.34 & 54.60 & 73.76 & 61.00 & 69.52 & 73.98 & \avgcell{68.35}
& 66.15 & 57.56
& 71.33 & 73.17 & \avgcell{72.25}
& 67.78 & 71.52 & 63.86 & \avgcell{67.69}
& 68.41 & 65.37 & \avgcell{67.76} \\

\rowcolor{groupgray}
\multicolumn{22}{l}{\textbf{\textit{White-box detectors: internal-state-based}}} \\

\rowcolor{wbrow}
EigenScore
& 54.78 & 50.87 & 72.95 & 53.28 & 73.41 & 55.67 & 64.85 & 70.32 & \avgcell{62.02}
& 56.45 & 50.76
& 66.23 & 78.04 & \avgcell{72.14}
& 51.25 & 69.57 & 74.70 & \avgcell{72.14}
& 62.19 & 53.00 & \avgcell{62.25} \\

\rowcolor{wbrow}
CCS
& 59.31 & 50.26 & 58.95 & 53.23 & 50.58 & 62.34 & 57.17 & 52.24 & \avgcell{55.51}
& 54.43 & 54.08
& 60.27 & 77.67 & \avgcell{68.97}
& 74.77 & 54.35 & 52.81 & \avgcell{53.58}
& 52.63 & 50.78 & \avgcell{57.40} \\

\rowcolor{wbrow}
HaloScope
& 67.98 & 62.30 & 56.30 & 59.52 & 62.17 & 61.12 & 60.27 & 54.91 & \avgcell{60.57}
& 61.28 & 51.67
& 60.74 & 67.15 & \avgcell{63.95}
& 58.90 & 74.35 & 55.70 & \avgcell{65.03}
& 62.05 & 67.17 & \avgcell{61.39} \\

\rowcolor{wbrow}
SAPLMA
& 81.27 & 72.12 & 91.72 & 76.89 & 88.67 & 71.72 & 80.27 & 85.61 & \avgcell{81.03}
& 77.98 & 66.60
& 83.01 & 83.62 & \avgcell{83.32}
& 69.87 & 81.30 & 61.17 & \avgcell{71.24}
& 93.15 & 68.22 & \avgcell{78.42} \\

\rowcolor{wbrow}
MIND
& 75.33 & 77.58 & 86.87 & 77.30 & 80.70 & 63.21 & 73.19 & 85.95 & \avgcell{77.52}
& 65.23 & 61.79
& 78.86 & 89.29 & \avgcell{84.08}
& 79.23 & 63.04 & 64.86 & \avgcell{63.95}
& 92.18 & 72.68 & \avgcell{75.72} \\

\rowcolor{wbrow}
SEP
& 81.95 & 69.31 & 89.98 & 76.84 & 88.25 & 76.32 & 79.05 & 86.26 & \avgcell{81.00}
& 74.46 & 64.80
& 78.62 & 86.46 & \avgcell{82.54}
& 76.24 & 74.35 & 62.46 & \avgcell{68.41}
& 87.82 & 69.76 & \avgcell{77.82} \\

\rowcolor{wbrow}
ICR Probe
& 68.16 & 65.24 & 62.90 & 59.88 & 53.61 & 58.27 & 52.23 & 64.87 & \avgcell{60.65}
& 57.34 & 58.14
& 54.11 & 59.33 & \avgcell{56.72}
& 73.90 & 65.22 & 65.43 & \avgcell{65.33}
& 61.36 & 58.57 & \avgcell{61.09} \\

\rowcolor{wbrow}
PRISM
& 77.46 & 73.03 & 91.25 & 75.88 & 90.18 & 79.91 & 80.96 & 87.16 & \avgcell{81.98}
& 80.21 & 67.65
& 77.24 & 84.88 & \avgcell{81.06}
& 80.51 & 66.96 & 68.54 & \avgcell{67.75}
& 87.47 & 61.38 & \avgcell{78.27} \\

\rowcolor{famavg}
\textit{Family Avg.}
& 70.78 & 65.09 & 76.37 & 66.60 & 73.45 & 66.07 & 68.50 & 73.42 & \avgcell{70.04}
& 65.92 & 59.44
& 69.89 & 78.31 & \avgcell{74.10}
& 70.58 & 68.64 & 63.21 & \avgcell{65.93}
& 74.86 & 62.70 & \avgcell{69.05} \\

\midrule

\modelrowall{\texttt{Qwen3-14B}}

\rowcolor{groupgray}
\multicolumn{22}{l}{\textbf{\textit{Black-box detectors: text-output-based}}} \\

\rowcolor{bbrow}
Verbalized Conf.
& 57.14 & 59.78 & 63.72 & 65.23 & 61.78 & 74.00 & 58.63 & 75.26 & \avgcell{64.44}
& 67.82 & 54.23
& 60.19 & 57.58 & \avgcell{58.88}
& 51.28 & 70.00 & 51.03 & \avgcell{60.52}
& 58.94 & 64.55 & \avgcell{61.83} \\

\rowcolor{bbrow}
SelfCheck-BERT
& 58.86 & 51.65 & 68.54 & 66.04 & 58.17 & 50.52 & 53.37 & 74.84 & \avgcell{60.25}
& 54.08 & 54.17
& 54.45 & 78.37 & \avgcell{66.41}
& 64.61 & 56.96 & 59.31 & \avgcell{58.14}
& 70.19 & 52.66 & \avgcell{60.40} \\

\rowcolor{bbrow}
SelfCheck-NLI
& 50.66 & 58.23 & 82.57 & 67.08 & 62.79 & 56.91 & 59.41 & 65.60 & \avgcell{62.91}
& 70.04 & 58.89
& 53.71 & 68.44 & \avgcell{61.08}
& 50.49 & 52.61 & 72.39 & \avgcell{62.50}
& 59.80 & 67.92 & \avgcell{62.21} \\

\rowcolor{bbrow}
Lexical Sim.
& 50.27 & 52.89 & 76.02 & 64.90 & 64.05 & 50.17 & 56.37 & 72.14 & \avgcell{60.85}
& 52.66 & 51.94
& 66.40 & 98.37 & \avgcell{82.39}
& 66.36 & 64.57 & 59.89 & \avgcell{62.23}
& 63.52 & 55.68 & \avgcell{62.72} \\

\rowcolor{famavg}
\textit{Family Avg.}
& 54.23 & 55.64 & 72.71 & 65.81 & 61.70 & 57.90 & 56.95 & 71.96 & \avgcell{62.11}
& 61.15 & 54.81
& 58.69 & 75.69 & \avgcell{67.19}
& 58.19 & 61.04 & 60.66 & \avgcell{60.85}
& 63.11 & 60.20 & \avgcell{61.79} \\

\rowcolor{groupgray}
\multicolumn{22}{l}{\textbf{\textit{Gray-box detectors: likelihood-based uncertainty}}} \\

\rowcolor{gbrow}
Perplexity
& 66.60 & 67.65 & 77.96 & 68.86 & 65.18 & 52.84 & 60.13 & 78.38 & \avgcell{67.20}
& 73.76 & 54.13
& 73.72 & 81.48 & \avgcell{77.60}
& 55.33 & 73.04 & 62.58 & \avgcell{67.81}
& 79.90 & 73.73 & \avgcell{68.55} \\

\rowcolor{gbrow}
Self-eval.
& 72.71 & 67.60 & 86.13 & 59.75 & 79.35 & 71.66 & 75.38 & 78.72 & \avgcell{73.91}
& 53.99 & 61.57
& 63.95 & 99.85 & \avgcell{81.90}
& 70.92 & 64.35 & 58.74 & \avgcell{61.55}
& 72.09 & 63.34 & \avgcell{70.59} \\

\rowcolor{gbrow}
LN-Entropy
& 52.65 & 54.53 & 76.59 & 65.31 & 64.37 & 54.53 & 59.38 & 72.83 & \avgcell{62.52}
& 73.76 & 50.97
& 66.09 & 98.96 & \avgcell{82.53}
& 69.35 & 61.52 & 54.74 & \avgcell{58.13}
& 68.80 & 55.69 & \avgcell{64.71} \\

\rowcolor{gbrow}
SAR
& 54.52 & 53.59 & 74.84 & 62.48 & 63.44 & 55.91 & 56.22 & 75.18 & \avgcell{62.02}
& 74.65 & 50.63
& 66.93 & 89.78 & \avgcell{78.36}
& 66.13 & 56.09 & 53.59 & \avgcell{54.84}
& 70.74 & 55.45 & \avgcell{63.54} \\

\rowcolor{gbrow}
Semantic Ent.
& 50.27 & 52.71 & 71.36 & 60.16 & 61.60 & 51.36 & 55.11 & 66.83 & \avgcell{58.68}
& 57.89 & 52.01
& 54.05 & 69.04 & \avgcell{61.55}
& 62.33 & 55.22 & 58.86 & \avgcell{57.04}
& 58.49 & 55.69 & \avgcell{58.41} \\

\rowcolor{famavg}
\textit{Family Avg.}
& 59.35 & 59.22 & 77.38 & 63.31 & 66.79 & 57.26 & 61.24 & 74.39 & \avgcell{64.87}
& 66.81 & 53.86
& 64.95 & 87.82 & \avgcell{76.39}
& 64.81 & 62.04 & 57.70 & \avgcell{59.87}
& 70.00 & 60.78 & \avgcell{65.16} \\

\rowcolor{groupgray}
\multicolumn{22}{l}{\textbf{\textit{White-box detectors: internal-state-based}}} \\

\rowcolor{wbrow}
EigenScore
& 54.77 & 50.47 & 74.27 & 60.50 & 63.36 & 53.42 & 57.50 & 64.42 & \avgcell{59.84}
& 53.12 & 50.06
& 67.67 & 96.30 & \avgcell{81.99}
& 66.00 & 54.35 & 53.10 & \avgcell{53.73}
& 68.15 & 54.48 & \avgcell{61.29} \\

\rowcolor{wbrow}
CCS
& 60.72 & 54.91 & 66.87 & 66.71 & 55.06 & 63.06 & 57.18 & 69.09 & \avgcell{61.70}
& 57.45 & 53.70
& 54.08 & 89.48 & \avgcell{71.78}
& 65.56 & 53.48 & 53.76 & \avgcell{53.62}
& 55.13 & 50.74 & \avgcell{60.41} \\

\rowcolor{wbrow}
HaloScope
& 55.19 & 64.94 & 54.01 & 56.08 & 50.12 & 53.03 & 58.74 & 61.03 & \avgcell{56.64}
& 57.45 & 52.45
& 55.87 & 83.85 & \avgcell{69.86}
& 60.60 & 50.00 & 51.88 & \avgcell{50.94}
& 57.11 & 62.37 & \avgcell{57.92} \\

\rowcolor{wbrow}
SAPLMA
& 69.43 & 65.70 & 86.55 & 75.09 & 84.22 & 70.28 & 73.01 & 67.52 & \avgcell{73.98}
& 76.42 & 64.95
& 60.14 & 94.22 & \avgcell{77.18}
& 72.30 & 56.09 & 67.08 & \avgcell{61.59}
& 87.97 & 74.92 & \avgcell{73.29} \\

\rowcolor{wbrow}
MIND
& 75.73 & 68.41 & 88.56 & 75.27 & 84.95 & 74.12 & 79.01 & 80.44 & \avgcell{78.31}
& 83.69 & 66.07
& 70.88 & 87.85 & \avgcell{79.37}
& 80.64 & 50.00 & 58.01 & \avgcell{54.01}
& 87.59 & 75.20 & \avgcell{75.67} \\

\rowcolor{wbrow}
SEP
& 54.71 & 66.84 & 86.31 & 78.98 & 85.75 & 67.59 & 72.29 & 70.71 & \avgcell{72.90}
& 53.90 & 65.09
& 57.19 & 59.41 & \avgcell{58.30}
& 63.81 & 74.78 & 63.97 & \avgcell{69.38}
& 83.13 & 71.30 & \avgcell{69.16} \\

\rowcolor{wbrow}
ICR Probe
& 51.76 & 67.31 & 65.03 & 67.41 & 54.88 & 63.68 & 57.64 & 54.79 & \avgcell{60.31}
& 60.12 & 51.96
& 65.19 & 76.59 & \avgcell{70.89}
& 55.16 & 52.17 & 53.79 & \avgcell{52.98}
& 50.36 & 60.83 & \avgcell{59.33} \\

\rowcolor{wbrow}
PRISM
& 58.76 & 69.55 & 87.55 & 75.87 & 85.38 & 55.26 & 58.67 & 76.71 & \avgcell{70.97}
& 68.97 & 66.50
& 53.82 & 89.93 & \avgcell{71.88}
& 72.09 & 57.39 & 55.31 & \avgcell{56.35}
& 82.62 & 57.37 & \avgcell{68.93} \\

\rowcolor{famavg}
\textit{Family Avg.}
& 60.13 & 63.52 & 76.14 & 69.49 & 70.47 & 62.56 & 64.26 & 68.09 & \avgcell{66.83}
& 63.89 & 58.85
& 60.61 & 84.70 & \avgcell{72.66}
& 67.02 & 56.03 & 57.11 & \avgcell{56.57}
& 71.51 & 63.40 & \avgcell{65.75} \\

\bottomrule
\end{tabular}%
}
\end{table}

\begin{table*}[!t]
\centering
\small
\caption{
Selected AUROC results (\%) on Llama-3.3-70B-Instruct.
We report selected experiments on seven datasets.
Higher values are better.
}
\label{tab:llama33_70b_selected_auroc}

\setlength{\tabcolsep}{4.6pt}
\renewcommand{\arraystretch}{1.05}

\begin{tabular}{@{}lrrrrrrr@{}}
\toprule
\rowcolor{modelgray}
\multicolumn{8}{@{}l}{\textbf{\texttt{Llama-3.3-70B-Instruct}}} \\
\textbf{Method} &
\textbf{HumanEval} &
\textbf{SQuAD} &
\textbf{ARC-C} &
\textbf{TruthfulQA} &
\textbf{MBPP} &
\textbf{HaluEval-QA} &
\textbf{TriviaQA} \\
\midrule

\rowcolor{modelgray}
\multicolumn{8}{@{}l}{\textit{\textbf{Black-box detectors: text-output-based}}} \\
\rowcolor{bbrow}
Verbalized Conf. & 68.08 & 70.52 & 68.28 & 73.18 & 69.00 & 65.36 & 80.41 \\
\rowcolor{bbrow}
SelfCheck-BERT & 52.31 & 62.13 & 71.12 & 66.99 & 64.43 & 77.94 & 75.17 \\
\rowcolor{bbrow}
SelfCheck-NLI & 61.15 & 57.10 & 66.22 & 62.83 & 67.79 & 64.38 & 72.00 \\
\rowcolor{bbrow}
Lexical Sim. & 56.92 & 65.28 & 57.26 & 65.34 & 55.21 & 76.07 & 74.28 \\
\rowcolor{famavg}
\textit{Family Avg.} & 59.61 & 63.76 & 65.72 & 67.09 & 64.11 & 70.94 & 75.46 \\

\rowcolor{modelgray}
\multicolumn{8}{@{}l}{\textit{\textbf{Gray-box detectors: likelihood-based uncertainty}}} \\
\rowcolor{gbrow}
Perplexity & 61.54 & 68.16 & 81.09 & 65.84 & 66.86 & 80.29 & 78.77 \\
\rowcolor{gbrow}
Self-eval. & 65.58 & 72.59 & 75.43 & 64.57 & 78.21 & 75.77 & 85.78 \\
\rowcolor{gbrow}
LN-Entropy & 50.77 & 63.72 & 57.11 & 62.23 & 65.71 & 76.77 & 76.03 \\
\rowcolor{gbrow}
SAR & 52.69 & 61.21 & 54.68 & 59.77 & 68.43 & 77.04 & 75.47 \\
\rowcolor{gbrow}
Semantic Ent. & 60.19 & 58.87 & 57.25 & 56.72 & 58.32 & 72.69 & 70.17 \\
\rowcolor{famavg}
\textit{Family Avg.} & 58.15 & 64.91 & 65.11 & 61.83 & 67.51 & 76.51 & 77.24 \\

\rowcolor{modelgray}
\multicolumn{8}{@{}l}{\textit{\textbf{White-box detectors: internal-state-based}}} \\
\rowcolor{wbrow}
EigenScore & 50.77 & 62.92 & 59.16 & 58.98 & 58.64 & 78.95 & 72.43 \\
\rowcolor{wbrow}
CCS & 65.38 & 53.22 & 63.10 & 52.29 & 56.64 & 72.90 & 56.05 \\
\rowcolor{wbrow}
HaloScope & 50.77 & 62.43 & 70.43 & 54.38 & 63.07 & 66.84 & 57.02 \\
\rowcolor{wbrow}
SAPLMA & 61.15 & 82.31 & 68.20 & 81.13 & 69.79 & 80.12 & 87.08 \\
\rowcolor{wbrow}
MIND & 68.85 & 85.23 & 77.39 & 83.41 & 77.29 & 78.89 & 86.46 \\
\rowcolor{wbrow}
SEP & 76.54 & 81.12 & 75.20 & 77.60 & 78.29 & 71.08 & 85.03 \\
\rowcolor{wbrow}
ICR Probe & 56.67 & 50.97 & 50.54 & 58.08 & 64.43 & 73.82 & 52.12 \\
\rowcolor{wbrow}
PRISM & 67.69 & 85.75 & 81.64 & 71.12 & 77.43 & 83.48 & 86.48 \\
\rowcolor{famavg}
\textit{Family Avg.} & 62.23 & 70.49 & 68.21 & 67.12 & 68.20 & 75.76 & 72.83 \\

\bottomrule
\end{tabular}
\end{table*}

\section{Cost Analysis}
\label{app:cost_analysis}

\textbf{Profiling setup.}
We profile detector cost on Llama-3.2-3B-Instruct using four representative datasets: ARC-Challenge, CoQA, GSM8K, and MBPP. 
Each profiling run uses up to 100 examples from the corresponding dataset, with the same train/validation/test split construction as the main benchmark pipeline. 
All runs are executed on a single NVIDIA H800 PCIe GPU with PyTorch 2.11.0 and CUDA 13.0. 
The maximum generation length is set to 2048 tokens, and hardware statistics are sampled every 5 seconds when the profiled stage is long enough for periodic sampling.

\textbf{Cost accounting.}
We separate detector cost into feature preparation, detector fitting, and detector inference. 
Feature preparation includes detector-specific evidence acquisition, such as stochastic generations, auxiliary self-evaluation outputs, token log-probabilities, hidden states, or QA-feature caches. 
Detector fitting is only applicable to methods that train detector-specific parameters. 
Detector inference measures the final scoring step on the test examples after the required artifacts have been prepared. 
Because different detectors require different forms of evidence, we report both wall-clock time and evidence-acquisition statistics.

\textbf{Reported quantities.}
Table~\ref{tab:cost_analysis_llama32_3b_runtime} reports average runtime statistics over the four profiled datasets. 
\textit{AUROC} is the corresponding average detection performance, included to contextualize the cost measurements. 
\textit{Cost@100} is the artifact-inclusive wall-clock time for processing up to 100 examples under the profiling setup. 
\textit{Score-only} is the detector-side time recorded in the isolated detector run, including detector-side feature use, fitting when applicable, and scoring. 
\textit{Infer.} reports the final detector scoring latency per example after the required artifacts are available. 
\textit{Extra gen. calls} and \textit{Extra gen. tokens} count only additional model generations beyond the primary response generation. 
Therefore, a value of \(0.0\) in these two columns does not mean zero computation; it means that the detector does not require additional generated responses beyond the original model output, although it may still use cached logits, hidden states, feature extraction, fitting, or detector-side scoring.

\begin{table}[H]
\centering
\small
\setlength{\tabcolsep}{4.2pt}
\renewcommand{\arraystretch}{1.08}
\caption{
Average detector runtime cost on Llama-3.2-3B-Instruct across ARC-Challenge, CoQA, GSM8K, and MBPP.
Cost@100 is the artifact-inclusive wall-clock time for up to 100 examples under the profiling setup.
Extra generation calls and tokens are counted per example beyond the primary response generation.
}
\label{tab:cost_analysis_llama32_3b_runtime}
\begin{tabular}{lrrrrrr}
\toprule
\textbf{Detector} &
\textbf{AUROC} &
\textbf{Cost@100} &
\textbf{Score-only} &
\textbf{Infer.} &
\textbf{Extra gen.} &
\textbf{Extra gen.} \\
&
&
\textbf{(s)} &
\textbf{(s)} &
\textbf{ms/ex.} &
\textbf{calls/ex.} &
\textbf{tokens/ex.} \\
\midrule
Perplexity & 77.3 & 4.5 & 4.5 & 0.06 & 0.0 & 0.0 \\
Verbalize & 57.7 & 10.9 & 10.9 & 0.05 & 1.0 & 3.2 \\
SAPLMA & 58.3 & 4.6 & 4.6 & 0.76 & 0.0 & 0.0 \\
PRISM & 68.2 & 4.8 & 4.8 & 0.93 & 0.0 & 0.0 \\
HaloScope & 68.5 & 4.8 & 4.8 & 1.93 & 0.0 & 0.0 \\
SAR & 84.8 & 5.1 & 5.1 & 55.89 & 5.0 & 224.3 \\
ICR Probe & 53.2 & 13.0 & 13.0 & 1.00 & 0.0 & 0.0 \\
CCS & 63.0 & 18.7 & 18.7 & 1.63 & 0.0 & 0.0 \\
MIND & 62.9 & 84.3 & 0.5 & 4.38 & 0.0 & 0.0 \\
Self-Evaluator & 74.8 & 173.9 & 173.9 & 0.20 & 1.0 & 68.3 \\
EigenScore & 73.1 & 433.7 & 433.7 & 7.54 & 5.0 & 224.3 \\
LN-Entropy & 81.5 & 435.3 & 435.3 & 0.03 & 5.0 & 224.3 \\
Lexical Similarity & 76.8 & 435.3 & 435.3 & 5.62 & 5.0 & 224.3 \\
SelfCheck-NLI & 77.0 & 441.1 & 441.1 & 50.81 & 5.0 & 224.3 \\
Semantic Entropy & 64.4 & 443.8 & 443.8 & 74.02 & 5.0 & 224.3 \\
SelfCheck-BERTScore & 79.2 & 444.0 & 444.0 & 103.96 & 5.0 & 224.3 \\
SEP-SE & 73.4 & 452.9 & 452.9 & 0.87 & 5.0 & 224.3 \\
\bottomrule
\end{tabular}

\vspace{0.3em}
\begin{minipage}{0.96\linewidth}
\footnotesize
\textit{Note.} Extra generation calls and tokens are counted beyond the primary response generation. 
A value of \(0.0\) in these columns does not indicate zero computation; these methods may still require cached logits, hidden states, feature extraction, detector fitting, or detector-side scoring. 
Score-only time is the detector-side time recorded in the isolated detector run; Cost@100 additionally accounts for the required cached artifacts under the profiling setup.
\end{minipage}
\end{table}

\textbf{Observations.}
The profiling results show that detector cost varies with the type of evidence required. 
Methods that reuse single-response likelihoods or cached internal features, such as Perplexity, SAPLMA, PRISM, and HaloScope, do not require extra generated responses and therefore have \(0.0\) extra generation calls and tokens. 
Prompt-based auxiliary methods require additional model calls: Verbalize uses one short auxiliary response per example, while Self-Evaluator uses one longer auxiliary response. 
Sample-based methods require five additional generations per example, which is reflected in the extra-generation columns. 
Among these methods, the final scoring overhead also differs: likelihood- or lexical-similarity-based scoring is relatively light, whereas NLI-, BERTScore-, and semantic-clustering-based scoring adds additional per-example computation.

\textbf{Interpretation.}
These results should be interpreted as controlled profiling measurements, not as hardware-independent runtime estimates. 
Absolute costs depend on the GPU, software stack, sequence lengths, batching behavior, and whether shared artifacts are recomputed or reused. 
The main practical implication is that detector comparisons should report not only accuracy, but also the evidence required to obtain the detector score.

\section{Assumptions and Limitations}
\label{app:assumptions_limitations}

OpenHalDet is designed as a unified evaluation benchmark for hallucination detection across heterogeneous generation scenarios. 
We summarize below the main assumptions behind the benchmark protocol and the scope of its limitations.

\subsection{Assumptions}
\label{app:assumptions}

\textbf{Reference-grounded annotation.}
OpenHalDet assumes that hallucination labels can be assigned by comparing a generated response against the available references, acceptable answers, task context, and known incorrect answers when provided. 
This enables a consistent annotation protocol across heterogeneous datasets, but the resulting labels necessarily depend on the coverage and quality of the available references. 
Responses that are correct but not captured by the provided references may be difficult to label automatically, and ambiguous or under-specified questions may introduce additional uncertainty. 
To reduce ambiguity, we use a structured LLM-judge annotation protocol and exclude abstention cases from the main binary evaluation. 
The resulting labels should therefore be interpreted as scalable reference-grounded labels rather than human-certified factuality judgments.

\textbf{Unified detector scoring.}
OpenHalDet evaluates each method as a scalar hallucination-risk detector, where larger scores indicate a higher likelihood of hallucination. 
This common interface is necessary for applying the same AUROC, AUPR, and FPR@95TPR metrics across black-box, gray-box, and white-box methods. 
For detectors whose raw score direction may differ across implementations or tasks, the reported metrics use the same score-orientation rule across methods, so that results measure separability under a shared evaluation convention.

\textbf{Shared evaluation protocol.}
The benchmark assumes that detector comparisons should be made under the same response generation, annotation, split construction, and metric implementation. 
Accordingly, detectors evaluated on the same dataset and backbone operate on the same generated responses and labels, and methods requiring additional evidence use cached artifacts produced by the benchmark pipeline. 
This design isolates detector behavior from protocol-level differences as much as possible.

\subsection{Limitations and Scope}
\label{app:limitations_scope}

\textbf{Benchmark-compatible implementations.}
Several prior detectors were originally developed under task-specific datasets, prompting formats, access assumptions, or computational budgets. 
When a method does not directly transfer to all OpenHalDet scenarios, we use a benchmark-compatible implementation that preserves the method's core input signal and scoring principle while making it compatible with the unified pipeline. 
For example, supervised probes are fitted on the OpenHalDet training split, methods requiring stochastic samples use the shared stochastic-generation artifacts, and methods requiring internal states use the shared hidden-state or QA-feature caches. 
For computationally heavier methods, we use scalable variants that are compatible with broad multi-dataset evaluation; for instance, our SAR implementation uses cached stochastic generations and sequence-level likelihood/similarity signals rather than requiring task-specific token-level processing. 
These choices support controlled comparison across tasks and backbones, but they may not exactly match every configuration used in the original source papers.

\textbf{Model and task coverage.}
OpenHalDet covers a broad set of task scenarios and evaluates detectors on recent open-weight LLMs from the Llama and Qwen families. 
However, the benchmark does not exhaust the space of possible deployment settings. 
Closed-source API models, domain-specialized models, multimodal generation, very long-context generation, and fully interactive multi-turn agent environments may exhibit different hallucination patterns and detector behavior. 
Similarly, although OpenHalDet includes multilingual and agentic scenarios, its coverage remains limited relative to the full diversity of languages, tools, and real-world user intents.

\textbf{Cost profiling scope.}
Our cost analysis is intended to compare relative cost patterns under a controlled hardware and software environment. 
Absolute runtime, memory, and energy values can vary with GPU type, system load, implementation details, sequence length, and whether shared artifacts are recomputed or reused. 
For this reason, we report artifact-inclusive cost together with explicit evidence-acquisition statistics such as extra model calls and generated tokens. 
The reported cost results should be interpreted as controlled profiling evidence rather than hardware-independent constants.

\textbf{Statistical uncertainty.}
We report bootstrap confidence intervals for representative AUROC comparisons to estimate finite-sample uncertainty over test examples. 
This procedure does not rerun response generation, annotation, detector training, or stochastic sampling. 
Thus, it captures uncertainty due to the finite test set, but not all sources of variation, such as random seeds, annotation variance, model sampling variance, or hardware-level nondeterminism. 
A more exhaustive uncertainty analysis would require repeated end-to-end runs, which is substantially more expensive for detectors requiring stochastic generations or hidden-state extraction.

\textbf{Intended use.}
OpenHalDet is intended for controlled comparison of hallucination detectors under shared generation, annotation, scoring, and evaluation protocols. 
It does not certify detectors for direct deployment in high-stakes applications, where additional task-specific validation, calibration, human review, and monitoring would be required.

\section{Artifact}
\label{app:artifact}

We provide a public repository accompanying this preprint.
The repository contains the core OpenHalDet codebase, including dataset adapters, prompt construction utilities, response-generation scripts, annotation scripts, detector implementations, and evaluation code.

\textbf{Repository contents.}
The artifact is intended to support inspection and reproduction of the benchmark pipeline. 
It includes code for converting datasets into the unified schema, generating model responses, running the LLM-judge annotation, preparing detector-specific features, and computing evaluation metrics. 
The repository also includes basic usage instructions and configuration examples.

\textbf{Data access.}
OpenHalDet builds on existing public datasets and benchmark sources. The repository provides adapters and processing scripts rather than claiming ownership of the original datasets. 
Users should follow the licenses and access terms of the corresponding source datasets.

\textbf{Compute requirements.}
Running the full benchmark can require substantial GPU resources, especially for detectors that use stochastic generations or hidden-state extraction. 
The code therefore supports running selected datasets, backbones, or detectors independently, and reusing cached intermediate artifacts when available.

\section{Statistical Significance}
\label{app:statistical_significance}

To provide a lightweight estimate of finite-sample uncertainty, we report 95\% stratified bootstrap confidence intervals for representative AUROC results on Llama-3.2-3B-Instruct. 
For each selected detector--dataset setting, we first fix the score orientation on the full test set using the same rule as the main AUROC evaluation, and then resample positive and negative test examples with replacement for 1,000 bootstrap trials. 
This procedure does not rerun response generation, annotation, detector fitting, or stochastic sampling; it estimates uncertainty due to the finite test set.

Table~\ref{tab:bootstrap_ci_llama32_3b} reports representative detectors from black-box, gray-box, and white-box access regimes across seven datasets. 
The intervals should be interpreted as finite-sample uncertainty estimates rather than exhaustive multi-seed variation or formal pairwise significance tests.

\begin{table*}[h]
\centering
\scriptsize
\setlength{\tabcolsep}{4.0pt}
\renewcommand{\arraystretch}{1.08}
\caption{
Representative 95\% stratified bootstrap confidence intervals for AUROC on Llama-3.2-3B-Instruct. 
Each cell reports AUROC followed by the 95\% confidence interval in brackets.
}
\label{tab:bootstrap_ci_llama32_3b}

\begin{tabular}{lcccc}
\toprule
\textbf{Detector} &
\textbf{TriviaQA} &
\textbf{TruthfulQA} &
\textbf{HotpotQA} &
\textbf{HaluEval-QA} \\
\midrule
\rowcolor{bbrow}
Verbalized Conf. & 74.27 [72.26, 76.34] & 59.49 [51.21, 66.89] & 63.64 [59.88, 67.26] & 68.13 [65.31, 70.86] \\
\rowcolor{bbrow}
SelfCheck-BERTScore & 82.67 [80.95, 84.49] & 59.54 [51.02, 68.11] & 70.58 [66.65, 74.26] & 80.03 [77.42, 82.45] \\
\rowcolor{bbrow}
SelfCheck-NLI & 84.52 [82.91, 86.27] & 66.26 [57.46, 75.60] & 70.68 [66.29, 74.69] & 71.30 [67.41, 75.10] \\
\rowcolor{bbrow}
Lexical Sim. & 82.50 [80.82, 84.33] & 61.79 [53.20, 70.41] & 71.96 [67.97, 75.92] & 78.92 [76.07, 81.75] \\
\rowcolor{gbrow}
Perplexity & 83.10 [81.34, 84.89] & 55.75 [46.36, 64.41] & 64.04 [59.96, 67.97] & 79.45 [77.06, 81.96] \\
\rowcolor{gbrow}
Self-Evaluation & 76.17 [74.07, 78.23] & 59.24 [50.29, 68.01] & 74.90 [71.06, 78.46] & 74.46 [71.32, 77.77] \\
\rowcolor{gbrow}
LN-Entropy & 81.35 [79.60, 83.09] & 50.88 [41.78, 59.59] & 70.81 [66.48, 74.64] & 79.13 [76.45, 81.90] \\
\rowcolor{wbrow}
SAPLMA & 82.69 [80.94, 84.47] & 75.15 [67.60, 82.01] & 77.51 [73.89, 80.92] & 81.85 [79.13, 84.42] \\
\rowcolor{wbrow}
MIND & 85.99 [84.55, 87.56] & 75.31 [67.46, 82.30] & 76.56 [72.66, 80.06] & 86.57 [83.96, 88.99] \\
\rowcolor{wbrow}
PRISM & 83.07 [81.39, 84.79] & 73.95 [66.14, 81.29] & 69.97 [65.79, 74.04] & 78.35 [75.48, 81.30] \\
\bottomrule
\end{tabular}

\vspace{0.7em}

\begin{tabular}{lccc}
\toprule
\textbf{Detector} &
\textbf{XSum} &
\textbf{SVAMP} &
\textbf{HumanEval} \\
\midrule
\rowcolor{bbrow}
Verbalized Conf. & 51.05 [46.20, 56.11] & 59.23 [51.79, 66.68] & 56.52 [50.00, 63.04] \\
\rowcolor{bbrow}
SelfCheck-BERTScore & 61.27 [53.25, 69.32] & 72.69 [62.44, 81.63] & 66.96 [46.52, 86.96] \\
\rowcolor{bbrow}
SelfCheck-NLI & 69.98 [62.73, 76.80] & 70.70 [58.89, 80.68] & 52.61 [30.87, 76.09] \\
\rowcolor{bbrow}
Lexical Sim. & 60.74 [52.94, 68.25] & 65.96 [53.63, 76.53] & 72.61 [53.91, 88.70] \\
\rowcolor{gbrow}
Perplexity & 57.91 [49.82, 66.00] & 64.76 [52.20, 76.88] & 52.61 [29.55, 77.39] \\
\rowcolor{gbrow}
Self-Evaluation & 59.34 [50.97, 67.04] & 79.74 [69.64, 88.89] & 86.74 [72.61, 96.96] \\
\rowcolor{gbrow}
LN-Entropy & 55.22 [47.24, 63.38] & 69.83 [57.07, 81.09] & 69.13 [51.73, 85.65] \\
\rowcolor{wbrow}
SAPLMA & 56.57 [48.41, 64.97] & 71.49 [59.92, 82.37] & 77.83 [56.08, 94.36] \\
\rowcolor{wbrow}
MIND & 59.41 [51.66, 66.97] & 79.72 [69.35, 88.58] & 67.83 [44.77, 86.52] \\
\rowcolor{wbrow}
PRISM & 62.41 [54.95, 70.07] & 73.02 [61.57, 82.88] & 63.91 [40.00, 83.92] \\
\bottomrule
\end{tabular}

\vspace{0.3em}
\begin{minipage}{0.96\textwidth}
\footnotesize
\textit{Note.} The number of evaluated examples is dataset-dependent: TriviaQA \(N=1964\), TruthfulQA \(N=154\), HotpotQA \(N=996\), HaluEval-QA \(N=1966\), XSum \(N=200\), SVAMP \(N=140\), and HumanEval \(N=33\). 
The wider intervals on smaller datasets, especially HumanEval, reflect larger finite-sample uncertainty.
\end{minipage}
\end{table*}

\section{Broader Impacts}
\label{app:broader_impacts}

OpenHalDet aims to support more transparent, reproducible, and comparable evaluation of hallucination detectors across diverse generation scenarios. 
By standardizing response generation, annotation, detector scoring, and metric computation, the benchmark provides a common protocol for analyzing the reliability and cost trade-offs of different detection methods.

At the same time, benchmark results should not be interpreted as certifying the factual reliability or deployment safety of LLM systems. 
Detector performance can depend on the covered datasets, backbone models, prompting protocols, automatic annotation quality, and evaluation metrics. 
Misuse could arise if benchmark scores are treated as standalone guarantees for high-stakes applications, or if automatic labels are assumed to be error-free. 
We therefore document the benchmark scope, annotation protocol, limitations, and intended use, and encourage users to complement OpenHalDet with task-specific validation and human review in safety-critical settings.

\section{Licenses}
\label{app:licenses}

OpenHalDet builds on existing open models and public benchmark datasets. 
We do not claim ownership of these third-party assets. 
All models and datasets should be used under the licenses and terms specified by their original providers. 
Table~\ref{tab:licenses} summarizes the licenses or usage terms associated with the main model and dataset assets used in our benchmark.

\begin{table}[H]
\centering
\scriptsize
\setlength{\tabcolsep}{4pt}
\renewcommand{\arraystretch}{1.08}
\caption{
Licenses and usage terms of the main third-party assets used in OpenHalDet. 
Users should refer to the original model and dataset sources for the authoritative license text.
}
\label{tab:licenses}
\begin{tabular}{lll p{6.2cm}}
\toprule
\textbf{Asset} & \textbf{Type} & \textbf{License / Terms} & \textbf{Notes} \\
\midrule
Llama-3.1-8B-Instruct & Model & Llama 3.1 Community License & Subject to Meta's Llama community license terms. \\
Llama-3.2-3B-Instruct & Model & Llama 3.2 Community License & Subject to Meta's Llama community license terms. \\
Llama-3.3-70B-Instruct & Model & Llama 3.3 Community License & Subject to Meta's Llama community license terms. \\
Qwen3-8B & Model & Apache 2.0 & Open-weight model from the Qwen family. \\
Qwen3-14B & Model & Apache 2.0 & Open-weight model from the Qwen family. \\
\midrule
ARC-Challenge & Dataset & Apache 2.0 & License stated by the corresponding source repository. \\
CommonsenseQA & Dataset & MIT & License stated by the corresponding source repository. \\
TriviaQA & Dataset & Apache 2.0 & License stated by the corresponding source repository. \\
TruthfulQA & Dataset & Apache 2.0 & License stated by the corresponding source repository. \\
SQuAD v2 & Dataset & CC BY-SA 4.0 & License stated by the corresponding dataset source. \\
HotpotQA & Dataset & CC BY-SA 4.0 & License stated by the corresponding dataset source. \\
CoQA & Dataset & Mixed license & Includes sources with different terms, such as CC BY-SA 4.0 and Apache 2.0. \\
HaluEval-QA & Dataset & MIT & License stated by the corresponding source repository. \\
RAGTruth & Dataset & MIT & License stated by the corresponding source repository. \\
XSum & Dataset & MIT & License stated by the corresponding source repository. \\
GSM8K & Dataset & MIT & License stated by the corresponding source repository. \\
SVAMP & Dataset & MIT & License stated by the corresponding dataset source. \\
TheoremQA & Dataset & MIT & License stated by the corresponding source repository. \\
xLAM-Agent & Dataset & Research-only terms & Used for academic/research evaluation. \\
Belebele & Dataset & CC-BY-NC 4.0 & Non-commercial license terms apply. \\
\bottomrule
\end{tabular}
\end{table}

The OpenHalDet repository contains dataset adapters and processing scripts rather than claiming ownership of the original datasets. 
When reconstructing the benchmark, users should download or access each dataset from its original source and comply with the corresponding license or usage terms. 
The code license for the OpenHalDet implementation is specified in the accompanying repository.

\section{Assets}
\label{app:assets}
\textbf{Released assets.}
The anonymized OpenHalDet release includes the benchmark codebase, dataset adapters, unified data schema, prompt and annotation templates, detector wrappers, evaluation scripts, and documentation.

\textbf{Documentation.}
The release includes a README and accompanying documentation describing installation, data preparation, schema fields, annotation protocol, detector evaluation, licenses, intended use, limitations, and reproduction commands.

\end{document}